\documentclass[10pt,a4paper]{article}
\usepackage[pdftex]{graphicx}
\usepackage[cmex10]{amsmath}
\usepackage{amsfonts}
\usepackage{amssymb}
\usepackage{amsthm}
\usepackage{siunitx}

\usepackage{algorithm}
\usepackage{algpseudocode}
\usepackage[hidelinks=true]{hyperref}

\usepackage{xfrac}
\usepackage{enumerate}

\usepackage[font=footnotesize ]{caption}
\usepackage{subcaption}

\usepackage{multirow}
\usepackage{rotating}
\usepackage{makecell}
\usepackage{booktabs}
\usepackage{tabularx,ragged2e}
\usepackage{longtable}
\usepackage{bm}
\usepackage{mathtools}
\usepackage{wrapfig}
\usepackage[mathscr]{euscript}
\usepackage{setspace}
\usepackage{rotating}

\usepackage{anysize}
\marginsize{2.5cm}{2.5cm}{2.5cm}{2.5cm}

\usepackage{xcolor}
\definecolor{correct}{rgb}{0.77, 0.12, 0.23} 
\definecolor{revise}{rgb}{0.0, 0.34, 0.25} 
\definecolor{insert}{rgb}{0.0, 0.2, 0.4} 
\definecolor{blue}{rgb}{0.0, 0.0, 0.0} 

\usepackage{parskip}
\theoremstyle{definition}
\newtheorem{definition}{Definition}

\theoremstyle{definition}

\usepackage{lipsum}

\usepackage{authblk}
\title{\textbf{Fragility, Robustness and Antifragility \\in Deep Learning}}
\author[1]{Chandresh Pravin}
\author[2]{Ivan Martino}
\author[3]{Giuseppe Nicosia}
\author[4]{Varun~Ojha\thanks{Corresponding Author: Varun~Ojha,~email: varun.ojha@ncl.ac.uk; \\Cite as: Pravin, Chandresh; Martino, Ivan, Nicosia, Giuseppe and Ojha, Varun (2023) \textit{Artificial Intelligence}, Elsevier}}

\affil[1]{University of Reading, United Kingdom}
\affil[2]{KTH Royal Institute of Technology, Sweden}
\affil[3]{University of Catania, Italy}
\affil[4]{Newcastle University, United Kingdom}

\date{}

\begin{document}

    \onehalfspacing
    \maketitle

\begin{abstract}

    We propose a systematic analysis of deep neural networks (DNNs) based on a signal processing technique for network parameter removal, in the form of \textit{synaptic filters} that identifies the \textit{fragility}, \textit{robustness} and \textit{antifragility} characteristics of DNN parameters. Our proposed analysis investigates if the DNN performance is impacted negatively, invariantly, or positively on both clean and adversarially perturbed test datasets when the DNN undergoes synaptic filtering.
    We define three \textit{filtering scores} for quantifying the fragility, robustness and antifragility characteristics of DNN parameters based on the performances for (i) clean dataset, (ii) adversarial dataset, and (iii) the difference in performances of clean and adversarial datasets. We validate the proposed systematic analysis on ResNet-18, ResNet-50, SqueezeNet-v1.1 and ShuffleNet V2 x1.0 network architectures for MNIST, CIFAR10 and Tiny ImageNet datasets. The filtering scores, for a given network architecture, identify network parameters that are \textit{invariant in characteristics} across different datasets over learning epochs. Vice-versa, for a given dataset, the filtering scores identify the parameters that are invariant in characteristics across different network architectures. We show that our synaptic filtering method improves the test accuracy of ResNet and ShuffleNet models on adversarial datasets when only the robust and antifragile parameters are selectively retrained at any given epoch, thus demonstrating applications of the proposed strategy in improving model robustness.  \\
    \textbf{Keyword:} Deep Neural Networks; Robustness Analysis; Adversarial Attacks; Parameter Filtering

\end{abstract}





%
%



\section{Introduction}
\label{sec:intro}
Deep neural networks (DNNs) are extensively used in various tasks and domains, achieving noteworthy performances in both research and real-world applications~\cite{lecun2015deep,samek2021explaining}. It is the critical weaknesses of DNNs, however, that warrant investigation if we are to better understand how they learn abstract relationships between inputs and outputs~\cite{papernot2016limitations,carliniTowards2017}. We propose to investigate the effects of a \textit{systematic analysis} on DNNs by using a signal processing technique for network parameter filtering (the terms DNN and network are used interchangeably), in contrast to random filtering~\cite{srivastava2014dropout,Yu_2018_CVPR,mariet2015diversity} methods. 

Our work analyzes the performance of a DNN under (a) \textit{internal stress} (i.e., the synaptic filtering of DNN parameters) and (b) \textit{external stress} (i.e., perturbations of inputs to the DNN). We define internal and external s
tress within the context of DNNs as a novel concept taking inspiration from the applications of stress on biological systems~\cite{oken2015systems}. Through analyzing the performance of a network to input perturbations (external stress) formed using an adversarial attack~\cite{szegedy2014intriguing,biggio2013evasion}, we bring the weakness of the DNN to the foreground. We simultaneously apply synaptic filtering (internal stress) to the network parameters in order to identify the specific parameters most susceptible to the input perturbations, thus characterizing them as \textit{fragile}. 
Similarly, we identify parameters of the DNN that are \textit{invariant} to both internal and external stress when considering the network performance, thus characterizing them as \textit{robust} to the applied stress. Following this reasoning, we introduce a novel notion of \textit{antifragility}~\cite{taleb2013mathematical} in deep learning as the circumstance in which any applied perturbations (internal and external) on a network result in an improvement of the network performance.

\begin{figure}[!t]
	\centering
    \includegraphics[width=0.97\textwidth]{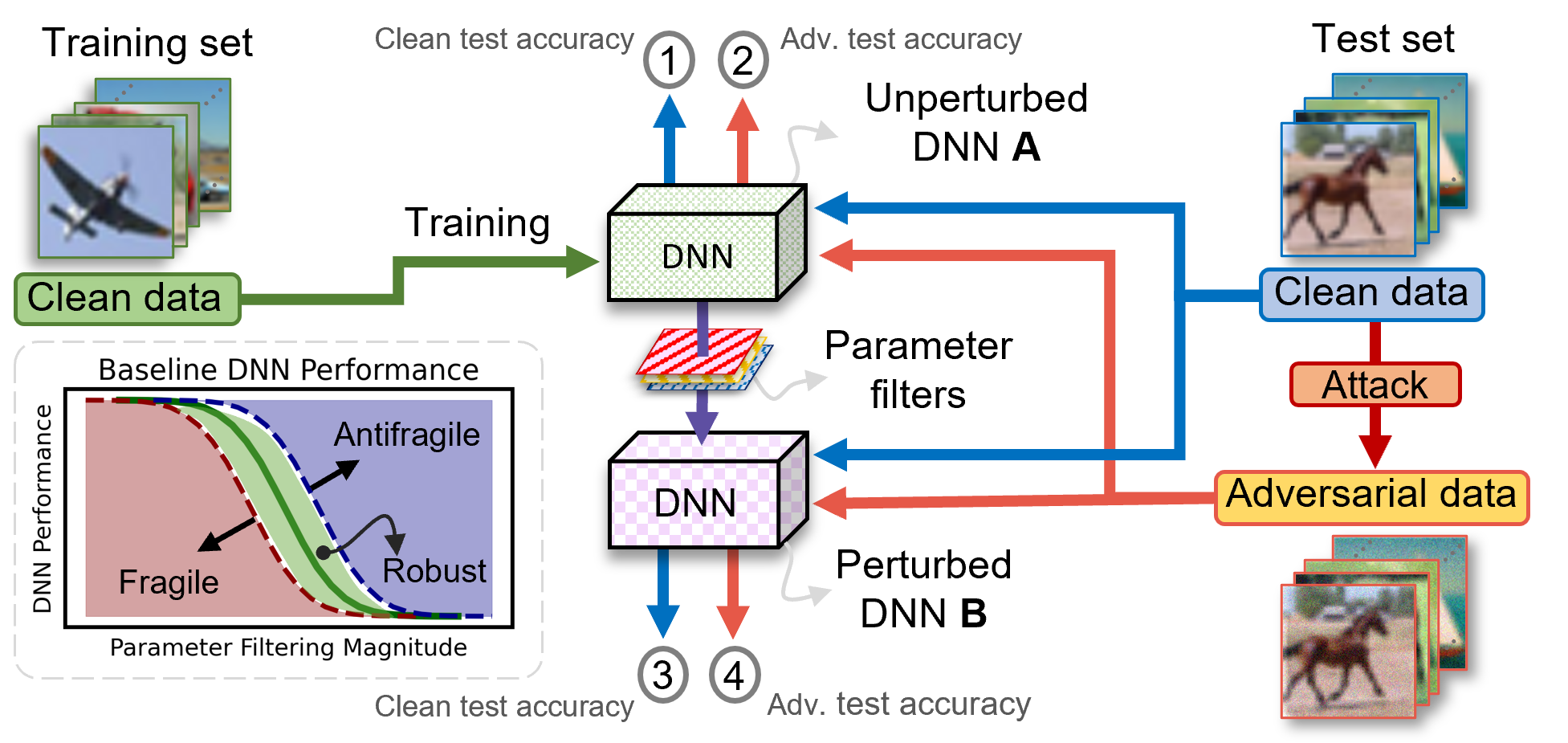}
        \caption{Our methodology of parameter filtering and evaluating DNN performances on clean and adversarial datasets. Passing a DNN through parameter filters is equivalent to internal stress and applying an adversarial attack with various magnitudes on clean data is equivalent to external stress on a DNN. In this methodology, the DNN performances (labeled 1, 2, 3, and 4) are individually compared against a defined baseline DNN performance (sold green line in the illustration shown on the lower left) in order to characterize DNN parameters as fragile (red shaded area), robust (green shaded area), or antifragile (blue shaded area).}
	  \label{fig:simple_intro}
\end{figure}

When considering external stress, such as variations to the network input, we focus our analysis specifically on varying magnitudes of adversarial attack perturbations~\cite{szegedy2014intriguing,biggio2013evasion} due to their ability to exploit the learned representations of a network to decrease network performance~\cite{freiesleben2022intriguing}. In our study, we focus on the fast gradient sign method (FGSM) attack for its equal single-step perturbation calculation for increasing network loss~\cite{goodfellow2015explaining}. Our synaptic filtering methodology (see Fig.~\ref{fig:simple_intro}) offers a comparative study of state-of-the-art DNNs using clean and adversarially perturbed datasets, and therefore, the study is relevant for any variation of perturbation introduced to the input space. We apply our methodology to expose the fragility, robustness and antifragility of network parameters over network learning epochs, which subsequently enables us to examine the \textit{landscape} (performance variations over epochs) of the network learning process.

In order to better understand how an adversarial attack is effective in bringing a network to failure~\cite{huang2017adversarial}, we take a novel methodology that considers network susceptibility to adversarial perturbations in conjunction with network architecture and the learning processes (see Fig.~\ref{fig:simple_intro}). The proposed synaptic filters are considered to be the \textit{lenses} under which we can \textit{characterize parameters} of network architecture. Introducing an adversarial attack to the methodology in Fig.~\ref{fig:simple_intro} offers a unique insight into how the characterization of network parameters varies between clean and adversarial inputs. We validated the analysis on the ResNet-18, ResNet-50~\cite{He_2016_CVPR}, SqueezeNet-v1.1~\cite{squeezenet2016iandola}, and ShuffleNet V2 x1.0~\cite{Ma_2018_ECCV} networks for the MNIST~\cite{lecun-mnisthandwrittendigit-2010}, CIFAR10~\cite{krizhevsky2009learning} and the Tiny ImageNet datasets~\cite{le2015tiny}. 

Our main contributions of this work, therefore, are as follows: 
\begin{itemize}
    \item We offer a novel methodology based on signal processing techniques that apply internal stress (parameter removal) and external stress (adversarial attack) on DNNs to characterize the network parameters as either fragile, robust, or antifragile.

    \item We offer parametric filtering scores that use a defined \textit{baseline network performance} to quantify the influence of specific parameters on the network performance. 

    \item We apply internal stress on networks in the form of synaptic filters and use the filtered network performances to show that networks trained on different datasets contain parameter characterizations that are \textit{invariant} to different datasets throughout the network training process. 
    
    \item We apply external stress to networks, in the form of an adversarial attack, to identify the \textit{specific parameters} targeted by the adversary through a comparison of the synaptic filtering performances of the clean and adversarial test datasets. 

    \item We show that our synaptic filtering method boosts the test accuracy of ResNet and ShuffleNet models on adversarial dataset when only the robust and antifragile parameters are retrained at any given epoch, thus proving a useful strategy for improving network robustness.
\end{itemize}

The following Sec.~\ref{sec:related_work} gives insights into the background and related works. Section~\ref{sec:prelim_defs} offers definitions of the terms and concepts introduced in the proposed methodology. Section~\ref{sec:synaptic_filtering} reports the proposed methodologies. Section~\ref{sec:res_anal} shows the experimental results acquired using the proposed methodologies, and Sec.~\ref{sec:conc} concludes the work.


\section{Background and related work}
\label{sec:related_work}
We propose evaluating the resilience of DNNs using a physiologically inspired approach concerning the resilience of humans to stress on their physiology~\cite{oken2015systems,karatsoreos2011psychobiological}. Therefore, we analyze the performance of DNNs to internal and external stress. Within the context of deep learning, we consider internal stress to be the perturbations to the network parameters (i.e., synaptic filtering)~\cite{ramanujan2020s,Molchanov_2019_CVPR} and we take external stress to be variations to the learning environment of the network (i.e., input perturbations)~\cite{gao2020fuzz,Wang2121demihuise,xu2020adversarial}. 

There exist various avenues of research when considering an analysis of DNNs to input  perturbations~\cite{goodfellow2015explaining,Tsipras2019RobustnessMB} and synaptic filtering~\cite{Molchanov_2019_CVPR,Yu_2018_CVPR}. The works of Szegedy et al.~\cite{szegedy2014intriguing} and Goodfellow et al.~\cite{goodfellow2015explaining} invited attention to investigate the vulnerability of DNNs to a particular method of crafting input perturbations in the form of adversarial attacks. The rapid development of new adversarial attacks~\cite{akhtar2018threat} and equally abundant adversarial defense  techniques~\cite{Samangouei2018DefenseGANPC,yuan2019adversarial}, call for methods of analyzing the resilience of DNNs to carefully crafted input perturbations, designed to bring networks to failure. 

The scrutiny of DNN resilience to these perturbations can be expanded to incorporate perturbations into network architectures. The study proposed by Han et al.~\cite{han2015learning} details how network parameters can be filtered out to reduce network size, without significantly affecting network performance. However, there may be conditions when filtering parameters may lead to improvements in the network performance. Therefore, we use a notion of antifragility to describe an increase in network performance whilst being subjected to internal and/or external stress, in the form of synaptic filters~\cite{mariet2015diversity,Molchanov_2019_CVPR,Yu_2018_CVPR} and adversarial attacks~\cite{huang2017adversarial,akhtar2018threat,yuan2019adversarial}. Our notion of antifragility in DNN is in line with the antifragility notion described by Taleb and Douady \cite{taleb2013mathematical}) to refer to a phenomenon whereby a system subjected to stress shows to improve in performance. We describe the related works on internal and external stress as follows:

\paragraph{Internal Stress (Parameter Filtering)} Network architecture affects how and what DNNs learn~\cite{sankararaman2020the,kornblith2019similarity,nakkiran2021deep,ojha2022backpropagation}. Therefore, the works of Ilyas et al.~\cite{ilyas2019adversarial} highlight the presence of robust and non-robust features within networks. In a similar context, we highlight the presence of fragile, robust and antifragile~\cite{taleb2013mathematical} parameters of different network architectures on both clean and adversarial test datasets. For the characterization of the network parameters, we propose a synaptic filtering methodology (see Fig.~\ref{fig:simple_intro}). 


Identifying fragile, robust and antifragile parameters informs us about the \textit{compressibility} of a  network based on the variation and degradation in the network performance~\cite{pravin2021adversarial}. A central principle of network compression techniques is to reduce network size whilst retaining network performance~\cite{han2015learning}. A method of achieving network compression is through using network pruning techniques~\cite{Davis2020What,Molchanov_2019_CVPR,Yu_2018_CVPR}. Our work of parameter filtering differs from the objective of pruning techniques that aim to reduce DNN size, whereas we aim to analyze the characteristics of DNN parameters by systematically filtering them. As well as our works differ from those systematic tuning of DNN hyperparameters such as the number of layers and number of neurons in a layer to analyze the DNN performance~\cite{taylor2021sensitivity}, i.e., we systematically internal architecture of the DNN. 

Siraj et al.~\cite{Siraj2019Robust} proposed a robust sparse regularisation method for network compactness while simultaneously optimizing network robustness to adversarial attacks. Similarly, we use our synaptic filtering methodology (a network parameter removal technique) to study the performances of a DNN on clean and adversarial datasets, which enable us to identify parameters that cause a decrease in network performance on the adversarial dataset~\cite{Ye_2019_ICCV} compared to the clean dataset, thus characterized as fragile in our work. We characterize parameters that are invariant to synaptic filtering on both clean and adversarial datasets as robust. Whereas the parameters that, when filtered, show to increase the network performance on the adversarial dataset compared to the clean dataset are characterized as antifragile.

\paragraph{External Stress (Adversarial Attacks)}
There are numerous methods for computing adversarial attacks on DNNs in the literature~\cite{Wang2121demihuise,xu2020adversarial}. The primary objective of adversarial attacks is to deceive a network into misclassifying otherwise correctly classified inputs~\cite{szegedy2014intriguing,goodfellow2015explaining}. The importance of the analysis of adversarial attacks on DNNs is significant due to the existence of adversarial examples in real world applications~\cite{kurakin2017adversarial,wang2021adversarial}. Similarly, in our work, we analyze the adversarial attack in order to characterize network parameters into the parameters that affect network performance negatively (fragile), invariantly (robust), and positively (antifragile).    
%
%
Adversarial examples are by design created to decrease network performance; however, when simultaneously carrying out synaptic filtering methods~\cite{Ye_2019_ICCV} it is possible to observe an increase in network performance, even under an adversarial attack, thus requiring the notion of antifragility.



\section{Definitions}
\label{sec:prelim_defs}
\begin{figure}[h!]
\centering
\begin{subfigure}{0.56\linewidth}
  \centering
  \includegraphics[width=1\textwidth]{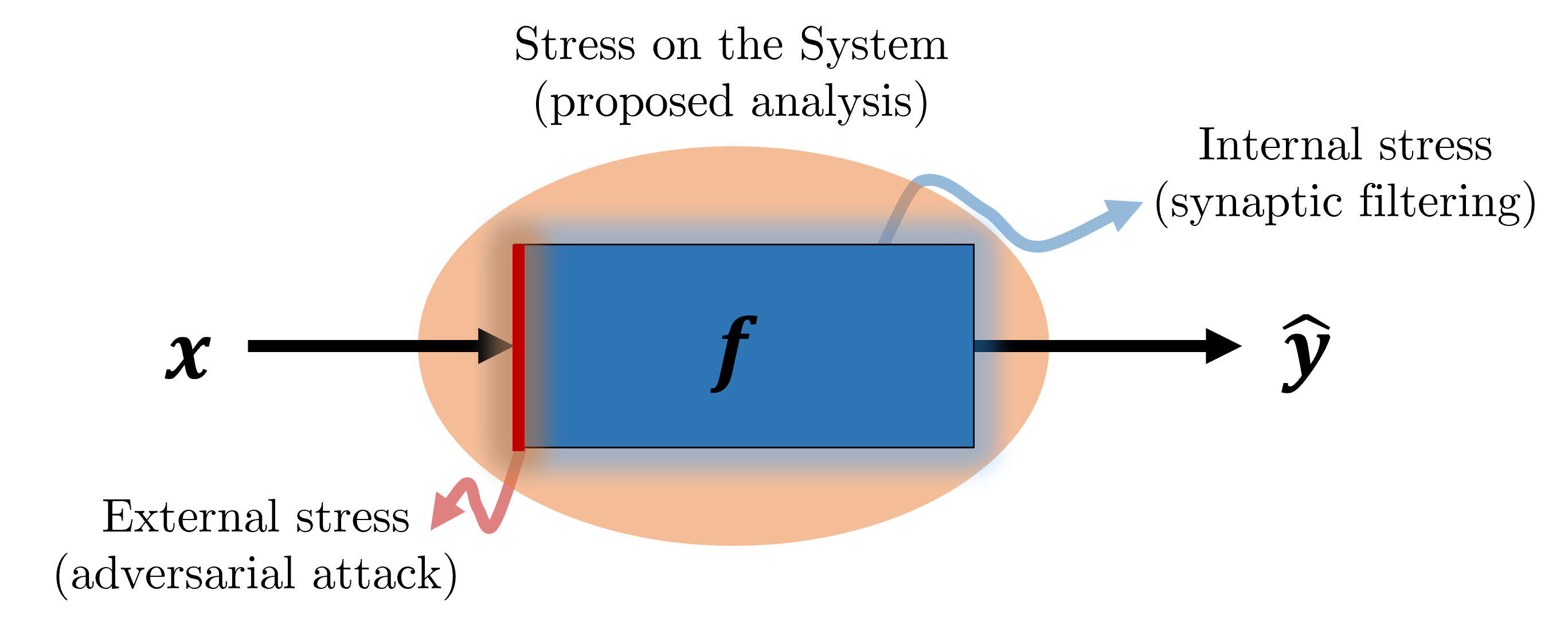}
  \caption{Evaluation overview}
  \label{subfig:local_sub1}
\end{subfigure}%
\begin{subfigure}{0.44\textwidth}
  \centering
    \includegraphics[width=0.75\textwidth]{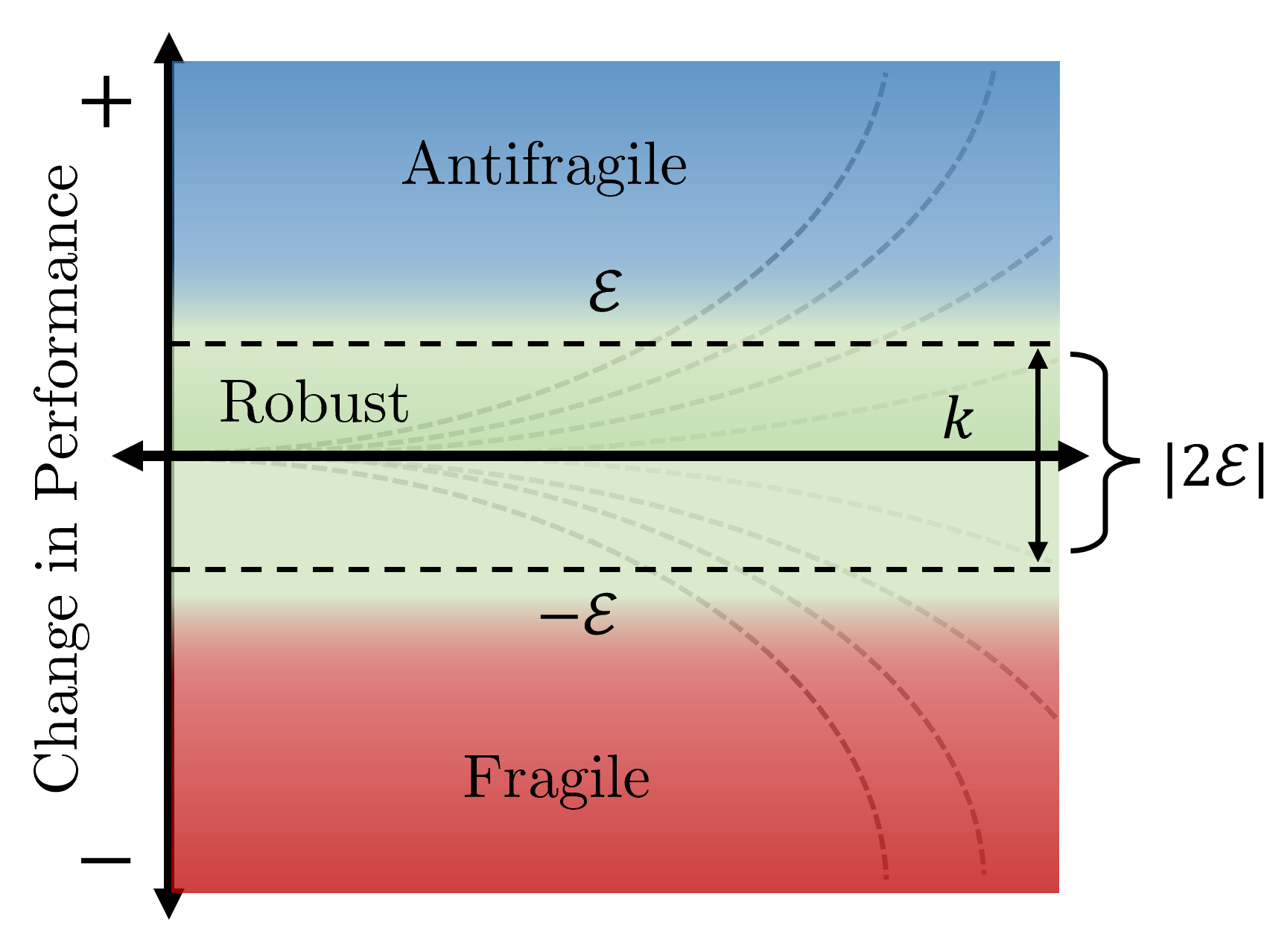}
  \caption{Response characteristics}
  \label{subfig:local_sub2}
\end{subfigure}
\caption{(a) Showing an overview of the proposed system evaluation method. (b) shows the characteristics of fragility, robustness and antifragility through analysing the performance of a system $\mathcal{F}$ whilst under stress.}
\label{fig:system_overview}
\end{figure}

In this Section, we define \textit{fragility}, \textit{robustness}, and \textit{antifragility} within the scope of DNNs. For defining fragility, robustness, and antifragility, we also need to define the \textit{internal stress}, \textit{external stress} and \textit{baseline network performance} of DNNs. Here the stress on a DNN is a systematic perturbation, either internal (synaptic filtering) or external (adversarial attack). The purpose of applying the stress on DNN is to test the operating conditions of the DNN for both learned and optimized states, when evaluated on unseen datasets. The concepts of network fragility, robustness, antifragility and stress are shown in Fig.~\ref{fig:system_overview}, where Fig.~\ref{subfig:local_sub1} shows the application of stress on a DNN and Fig.~\ref{subfig:local_sub2} shows the interpretation of DNN performance for parameter characterization. For detailed definitions of the above-mentioned concepts, we consider the following notations.

Consider a neural network \textit{architecture} as a set of functions $f(x,\cdot)$ that consists of a configuration of parameters, such as convolutions, batch normalization, pooling layers, activation functions, etc.~\cite{Davis2020What}, we define a \textit{parameterized} neural network as $f(x, \mathbf{W})$, for specific parameters $\mathbf{W}$ and input $x$. For an $l$-layer network with a $d$ dimensional input $x \in \mathbb{R}^d$; the $K$-class classification function is thus $f:\mathbb{R}^d \rightarrow \mathbb{R}^K$. The prediction of $f(x,\mathbf{W})$ is given by $\hat{y} = \arg\max_{1\leq k \leq K} f_k(x,\mathbf{W})$. The network parameters $\mathbf{W}$ are assumed to be optimized, either partially or fully, using back-propagation and a loss function $\mathcal{L}:\mathbb{R}\times\mathbb{R}\rightarrow\mathbb{R}$ given by $\mathcal{L}(\hat{{y}}, y)$ to calculate the network error. 


\subsection{Stress on DNNs}
\label{subsec:sys_under_stress}
To formulate internal stress on the network, we consider two filtering domains: \textit{local} (the parameters of any specific layer) and \textit{global} (the parameters of the whole network). We apply synaptic filtering to filter the parameters of \textit{trainable convolutional layers} and \textit{fully connected layers} of the network, the non-trainable parameters, however, remain unaffected by the synaptic filtering procedure. The $l$-th layer network parameters (local parameters) are given as $\mathbf{W}^{(l)}$, while the global network parameters are $\mathbf{W}$. For convenience, we denote the network parameters to be evaluated by the synaptic filtering methods as $\theta$, where $\theta = \mathbf{W}^{(l)}$ is the local parameter analysis~\cite{han2015learning} and $\theta = \mathbf{W}$ is the global parameter analysis, as mentioned in~\cite{Frankle2018Machine}. 

\begin{definition}[Synaptic filtering]
\label{def:synap_filt_def}
    The synaptic filtering involves taking a network $f(x,\theta)$ with parameter $\theta$ as an input and producing a filtered network $f(x, \tilde{\theta}_{\alpha})$ with filtered parameter $\tilde{\theta}_{\alpha})$ as:
    \begin{equation}
    \label{eq:param_filter}
    \begin{gathered}
       \tilde{\theta}_{\alpha} = B_{\alpha} \odot \theta_{\alpha}, \quad
        B_{\alpha} \in \{0,1\}^{|\theta_{\alpha}|},  
    \end{gathered}
    \end{equation}
    where {\color{blue}$\alpha=\{\alpha_{0}, \alpha_{1}, \dots,\alpha_{A}\}$ is the normalised synaptic filtering thresholds across the complete parameter range of the evaluated network/layer with a lower bound $\alpha_{0}=0$, upper bound $\alpha_{A}=1$ and step size $\Delta_\alpha$ given by $\alpha_{1} = \alpha_{0} + \Delta_\alpha$}. For synaptic filtering of a network, we have $\hat{y} = f(x,\theta)$ as the network predictions for the unperturbed network and $\hat{y}_{\alpha} = f(x,\tilde{\theta}_{\alpha})$ as the network predictions for the perturbed network. In Eq.~\ref{eq:param_filter}, $B_{\alpha}$ is a binary mask for a threshold $\alpha$ that filters parameters, $\theta_{\alpha}$ are the set of parameters to be filtered that may be different to $\theta$, and $\odot$ is the element-wise product operator
\end{definition}

To further constrain the internal stress analysis, we define that the network parameters to be filtered $\theta$ is not a zero vector prior to the synaptic filtering, i.e., $\theta$ must be a trained network: %
%
    $\theta \neq \textbf{0}$.
%
If this constraint is not met, the prediction of the network $f(x,\theta)$ will result in output values {\color{blue}of} zero for all inputs.  
\begin{definition}[Internal stress - synaptic filtering of the DNN parameters.]
\label{def:internal_stress}
The internal stress on a DNN is the application of the synaptic filtering method with various magnitudes of $\alpha$ ranging from a minimum filtering threshold $\alpha_{0}$ to the maximum filtering threshold $\alpha_{A}$ in order to obtain a set of $|\alpha|$ filtered networks $\mathcal{S}_{\alpha}$:

\begin{equation}
\label{eq_1:stress}
\begin{gathered}
        \mathcal{S}_{\alpha} = \{f(x,\tilde{\theta}_{\alpha_{0}}), f(x,\tilde{\theta}_{\alpha_{1}}),\dots, f(x, \tilde{\theta}_{\alpha_{A}})\}.
\end{gathered}
\end{equation}
\end{definition}

With evaluating a network to internal stress, we examine how the filtering of learned parameters of a network, influences the network performance, thus identifying the specific \textit{filtering thresholds} required to bring the network to failure.

Considering external stress as variations to the input $x$, we introduce $x_{\epsilon} = x+\delta_{\epsilon}$ as the \textit{perturbed example} of $x$ with an adversarial perturbation $\delta_{\epsilon} \in \mathbb{R}^d$, where $\epsilon=\{\epsilon_{0},\epsilon_{1}, \dots,\epsilon_{E}\}$ is the perturbation magnitude with minimum  perturbation magnitude $\epsilon_{0}$ and maximum perturbation magnitude $\epsilon_{E}$ with step size $\epsilon_{1} = \epsilon_{0} + \Delta_\epsilon$. Using a single adversarial attack formulation method $\delta$ we define $\hat{y} = f(x,\theta)$ as the network predictions on clean dataset and $\hat{y}_{\epsilon} = f(x_{\epsilon},\theta)$ as the network predictions on the adversarial dataset [Fig.~\ref{fig:simple_intro}(Left)]. When dealing with external stress only, $\theta$ is taken as the complete set of network parameters $\mathbf{W}$. 

The performance of $f(x_{\epsilon},\theta)$ can inform us of the ability of the network to remain stable to external stress (input perturbations) applied to the network. This is achieved through a comparison of the network performance on a clean dataset and an adversarially perturbed dataset. There are numerous variations of $\delta$ that can be used to form external stress to the network, from targeting specific features of $x$ to drawing distortions from a different distribution~\cite{Wang2121demihuise,xu2020adversarial}. However, in this work, we only focus on one perturbation method $\delta$, i.e., FGSM attack, as our objective is to only compare DNN performance on clean and perturbed inputs (Fig.~\ref{fig:simple_intro}). 

When applying external stress with various magnitude $\epsilon$, we get a set of perturbed inputs for the network $\mathcal{S}_{\epsilon}$: 

\begin{definition}[External stress - adversarial attack on DNN]
\label{def:external_stress}
The external stress on a DNN is the application of an adversarial attack with various perturbation magnitudes $\epsilon$ ranging from a minimum perturbation magnitude $\epsilon_{0}$ to the maximum perturbation magnitude $\epsilon_{E}$ in order to obtain a set of $|\epsilon|$ inputs to the network $\mathcal{S}_{\epsilon}$:
\begin{equation}
\label{eq_2:stress}
        \mathcal{S}_{\epsilon} = [f(x_{\epsilon_{0}}, \theta), \dots, f(x_{\epsilon_{E}}, \theta)].
\begin{gathered}
\end{gathered}
\end{equation}
\end{definition}

With external stress on a network we examine how the variations in the input environment influence the network performance, thus identifying the specific magnitudes of the attack required to bring the network to failure.




An important consideration to make when analyzing networks using internal and external stress in Definitions~\ref{def:internal_stress} and \ref{def:external_stress}, is that a resultant perturbed network  ($\mathcal{S}_{\alpha}$ and $\mathcal{S}_{\epsilon}$) may offer equal performance to the unperturbed network, i.e., for all inputs $x$ in test set, we observe:
\begin{gather*}
    ~~p(\hat{y}_\alpha=y|f(x,\tilde{\theta}_{\alpha})) \approx p(\hat{y}_\alpha=y|f(x, \theta)) \text{ for internal stress threshold } \alpha, \text{and}    \\
    p(\hat{y}_{\epsilon}=y|f(x_{\epsilon}, \theta)) \approx p(\hat{y}_{\epsilon}=y|f(x,\theta)) \text{ for external stress magnitude } \epsilon,
\end{gather*}
%

where $p(\cdot)$ is a function that measures the network accuracy over all inputs $x$. This indicates that even under stress, a DNN may perform equivalently to an unperturbed network. Therefore, in order to evaluate the performance of a network to stress, we must define a baseline network performance against which we can measure the performance of perturbed and unperturbed networks. 


A baseline network performance can vary for different types of stress (internal or external), as there may arise instances where the response of the baseline network performance, defined as $\hat{f}(x,\theta_{\alpha})$ for internal stress and $\hat{f}(x_{\epsilon},\theta)$ for external stress, is not necessarily the same as the performance of the initially trained network (unperturbed network) $f(x,\theta)$. The baseline network performance for a combination of internal and external stress is defined as $\hat{f}(x_{\epsilon},\theta_{\alpha})$, where the baseline network is a function of $\epsilon$ and $\alpha$.   

To give context on why baseline network performance may not necessarily be the same as the performance of an unperturbed network, take the example when we apply internal stress to a DNN, the result is a set of 
filtered networks $\mathcal{S}_{\alpha}$. If we define the upper bound of the stress magnitude equal to the total number of network parameters, i.e., $\alpha_{A} = |\theta|$, then we obtain a network with parameter value zero $\tilde{\theta}_{\alpha_{A}} = \textbf{0}$. Noticeably, the performance of a maximally perturbed network $f(x,\tilde{\theta}_{\alpha_{A}})$ cannot equal to the performance of unperturbed network $f(x, \theta)$, i.e., $f(x,\tilde{\theta}_{\alpha_{A}}) \neq f(x, \theta)$. Thus we require the baseline network performance to be a function of the magnitude of stress applied on the DNN.  A detailed description of baseline network performance is given later in Sec.~\ref{subsec:base_resp}. 


\subsection{Fragility, Robustness and Antifragility}
\label{subsec:frag_rob_antifrag}
Here we define the three characterizations of network parameters: fragility, robustness and antifragility. In order to define the different characterizations of network parameters, we must establish the stress to which we can evaluate network parameter fragility, robustness and antifragility. The stress in question may be internal ($\mathcal{S}_{\alpha}$) or external
($\mathcal{S}_{\epsilon}$), or a combination of the two. 

For simplicity, we consider only internal network stress for the definitions provided below. However, the change of variables from $\mathcal{S}_{\alpha}$ to $\mathcal{S}_\epsilon$, from $\hat{f}(x,\tilde{\theta}_{\alpha})$ to $\hat{f}(x_{\epsilon}, \theta)$, and from $\Delta_{\alpha}$ to $\Delta_\epsilon$ will give the definition of fragility, robustness and antifragility for external stress.

\begin{definition}[Fragility]
\label{def:frag_def}
The parameters of a network are fragile if the performance of the networks \textit{decreases} below a threshold $-\varepsilon$, compared to the baseline network performance for all magnitudes of the applied stress. Formally, the fragility to internal stress can be defined as:
%
\begin{equation}
\label{eq_1:fragility}
    \color{blue}{\sum_{i=0}^{A} [\mathcal{S}_{\alpha_{i}}-\hat{f}(x, \tilde{\theta}_{\alpha_{i}})] \Delta_{\alpha}< -\varepsilon},
\end{equation}
where $\Delta_{\alpha}$ is the change in synaptic filtering threshold $\alpha$, {\color{blue}$A$ is equal to $|\alpha|$}, $\varepsilon \geq 0$ and asserts a variable fragility measure, as shown in Fig.~\ref{subfig:local_sub2} (red shaded region). When the threshold $\varepsilon=0$, we have a strict fragility condition. Equation~\eqref{eq_1:fragility} computes the discrete area difference between the stressed network performance and the baseline network performance for all stress magnitudes of $\alpha$.
\end{definition}

\begin{definition}[Robustness]
\label{def:rob_def}
The parameters of a network are robust if the performance of the networks is \textit{invariant} to a threshold $\pm\varepsilon$, compared to the baseline network performance for all magnitudes of the applied stress. Formally, the robustness to internal stress can be defined as:
%
\begin{equation}
\label{eq_1:robustness}
    \color{blue} {-\varepsilon  \leq \sum_{i=0}^{A} [\mathcal{S}_{\alpha_{i}} - \hat{f}(x, \tilde{\theta}_{\alpha_{i}})] \Delta_{\alpha} \leq \varepsilon},
\end{equation}
where $\Delta_{\alpha}$ is the change in synaptic filtering threshold $\alpha$, $\varepsilon \geq 0$ and asserts a variable robustness measure, as shown in Fig.~\ref{subfig:local_sub2} (green shaded region). When the threshold $\varepsilon=0$, we have a strict robustness condition. Equation~\eqref{eq_1:fragility} computes the discrete area difference between the stressed network performance and the baseline network performance for all stress magnitudes of $\alpha$.
\end{definition}

\begin{definition}[Antifragility]
\label{subsubsec:antifrag_def}
The parameters of a network are antifragile if the performance of the networks \textit{increases} to a threshold $\varepsilon$, compared to the baseline network performance for all magnitudes of the applied stress. Formally, the antifragility to internal stress can be defined as:
%
\begin{equation}
\label{eq_1:antifragile}
    \color{blue} {\varepsilon  < \sum_{i=0}^{A} [\mathcal{S}_{\alpha_{i}} - \hat{f}(x, \tilde{\theta}_{\alpha_{i}})] \Delta_{\alpha}},
\end{equation}
where $\Delta_{\alpha}$ is the change in synaptic filtering threshold $\alpha$, $\varepsilon \geq 0$ and asserts a variable robustness measure, as shown in Fig.~\ref{subfig:local_sub2} (blue shaded region). When the threshold $\varepsilon=0$, we have a strict antifragility condition. Equation~\eqref{eq_1:fragility} computes the discrete area difference between the stressed network performance and the baseline network performance for all stress magnitudes of $\alpha$.
\end{definition}

\section{Methodology of DNN parameters characterization}
\label{sec:synaptic_filtering}
In this Section, we present the methodology of DNN parameter characterization that is shown in Fig.~\ref{fig:simple_intro}.  Concisely, Fig.~\ref{fig:simple_intro} shows that this methodology has two major aspects a) the application of internal and external stress on DNN in terms of synaptic filtering and adversarial attack, and b) the need of a process to characterize parameters into fragile, robust and antifragile. This section first explains how we apply internal and external stress on DNNs in Sec.~\ref{subsec:pahse_1} and then introduces parameter scores that characterize the parameters in Sec.~\ref{subsec:pahse_2}. Finally, we discuss the experiment setting in Sec.~\ref{subsec:exp_setup}.

\subsection{Framework of internal and external stress on DNNs}
\label{subsec:pahse_1}
\begin{figure}[!t]
	\centering
    \includegraphics[width=0.98\textwidth]
        {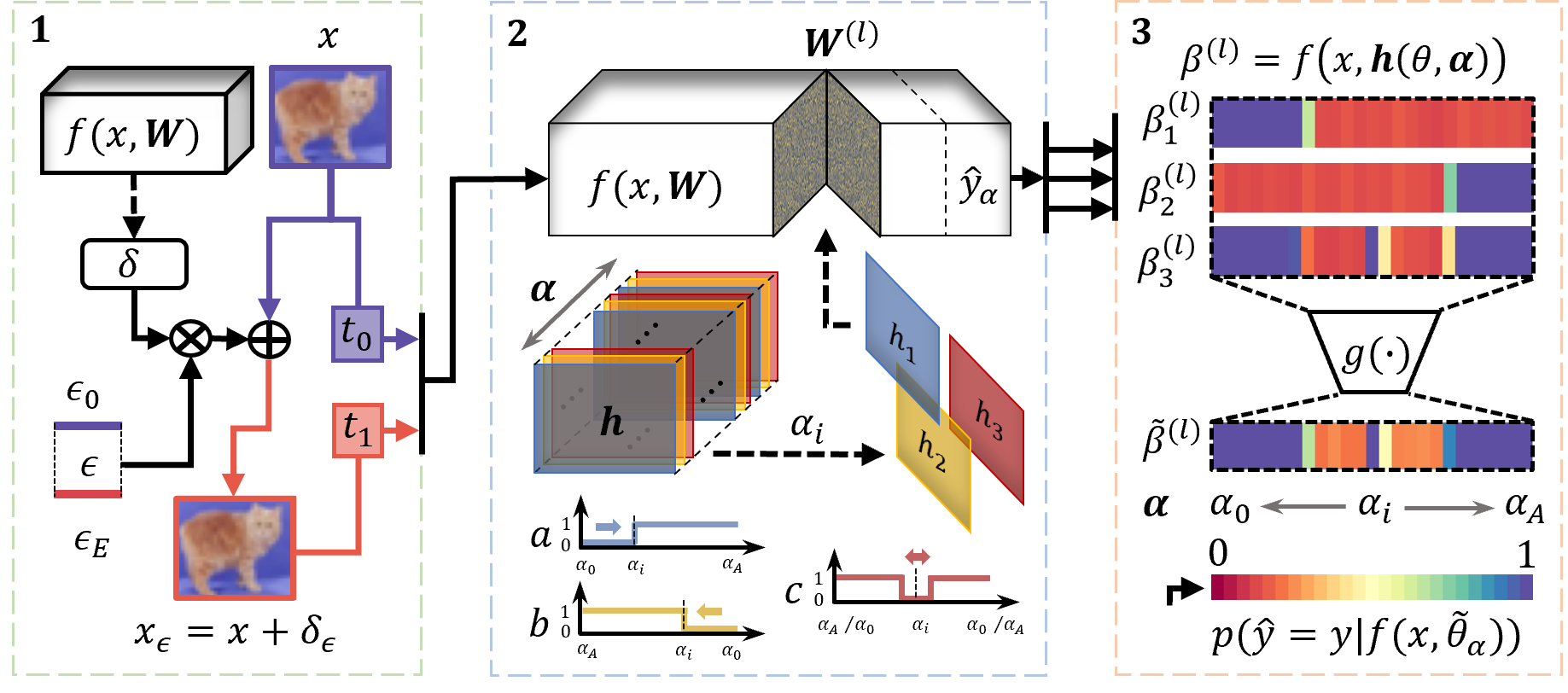}
	  \caption{synaptic filtering framework. \textit{Left} block (\textbf{1}) shows the input $x$ at time $t_0$; network $f(x,\textbf{\textit{W}}$) with parameters $\textbf{\textit{W}}$; the adversarial attack $\delta$ [this study computes $\delta$ using $f(x,\textbf{\textit{W}})$]; the perturbation magnitude $\epsilon$ and the resultant adversarial example $x_{\epsilon}$ at time $t_{1}$. The perturbation magnitude $\epsilon$ is bounded by ($\hat{y}_{\epsilon} \approx \hat{y}$) and ($\hat{y}_{\epsilon}{K} > 1$) for $K$ classes; $\hat{y}$ and $\hat{y}_{\epsilon}$ are clean and adversarial accuracies. \textit{Middle} block (\textbf{2}) outlines the set of synaptic filters $\textbf{\textit{h}}$, containing $h_{1}, h_{2}$ and $h_{3}$ filters at each point $\alpha_{i}$ applied to layer $\textbf{\textit{W}}^{(l)}$, resulting in the network performance to the filters. There are $\color{blue}{A}$ sets of $\textbf{\textit{h}}$ for each $\alpha_i \in [\alpha_0$, $\alpha_{A}]$. \textit{Right} block (\textbf{3}) shows $\beta^{(l)} = f(x,\textbf{\textit{h}}(\theta, \alpha))$ as the system performances for all values of $\alpha$, where $\theta$ is $\textbf{\textit{W}}^{(l)}$ for a local analysis at layer $l$. The function $g(\cdot)$ combines $\beta^{(l)}_{1}, \beta^{(l)}_{2}$ and $\beta^{(l)}_{3}$ into a combined system performance $\hat{\bar{\beta}}^{(l)}$.}
	  \label{fig:full_system}
\end{figure}

We systematically apply internal and external stress on DNNs. The process of internal and external stress on DNNs is shown in Fig.~\ref{fig:full_system}, which is a three-step framework (adversarial attack on DNNs, synaptic filtering of DNNs, combined network performance) that leads to parameter score calculation for the DNN parameter characterization.   

\subsubsection{Attack on DNNs}
In evaluating networks to internal stress, we compare the network performances to the synaptic filtering procedure for clean and adversarial (external stress) datasets (discussed in the following Sec.~\ref{subsec:synap_filt}). In this study, we work primarily with the FGSM attack~\cite{goodfellow2015explaining} for the adversarial perturbation formulation; other attack formulation methods would not affect the synaptic filtering described in this section. The synaptic filtering technique is designed to be applied to a network with any variation on the inputs, therefore, the nature of the attack formulation method can be changed without affecting the synaptic filtering technique.

In order to experiment with an adversarial dataset, we must define some constraints of the attack [Fig.~\ref{fig:full_system}(Left block)], such that the synaptic filtering responses are comparable between different network architectures and datasets. The constraints imposed upon the adversarial attack magnitude $\epsilon$ are, as follow:

\begin{definition}[minimum attack bound $\epsilon_{0}$ -- \textit{constraint 1}.]\label{con:min_attack} We limit the adversarial attack to follow $p(\hat{y}_{\epsilon}=y|x+\delta_{\epsilon_{0}}) < p(\hat{y}=y|x)$, for all inputs $x$ in the test dataset. This constraint allows us to select a suitable minimum attack magnitude $\epsilon_{0}$, such that otherwise correctly classified inputs are misclassified, due to the adversarial attack.
\end{definition}

\begin{definition}[\text{maximum attack bound} $\epsilon_{E}$ -- \textit{constraint 2}.]
\label{con:max_attack}  We limit the adversarial attack to a suitable maximum attack magnitude $\epsilon_{E}$, such that the network test accuracy is above a random guess ($\hat{y}_{\epsilon_{E}}K > 1 $), i.e., we have the constraint: $p(\hat{y}_{\epsilon_{E}}=y | x+\delta_{\epsilon_{E}})K > 1$, for all inputs $x$ in the test dataset.
\end{definition}

\begin{definition}[relative attack $\epsilon$ -- \textit{constraint 3}.]
\label{con:rel_attack}  To compare the performance of different network architectures and datasets to the synaptic filtering procedure, we must consider values of $\epsilon$ for different networks/datasets that reduce the network performance equally. Considering two different networks $f_{1}$ and $f_{2}$, we use a single attack $\delta$, for which $\epsilon_{1}$ and $\epsilon_{2}$ are the \textit{relative attack} magnitudes for $f_{1}$ and $f_{2}$. Suitable values of $\epsilon_{1}$ and $\epsilon_{2}$ should be chosen, such that $f_{1}(x, \theta) - f_{1}(x+\delta_{\epsilon_{1}}, \theta) \approx f_{2}(x, \theta) - f_{2}(x+\delta_{\epsilon_{2}}, \theta)$ thus ensuring that the adversarial perturbations affect the network performances equally.
\end{definition}

\subsubsection{Synaptic filtering of DNNs} 
\label{subsec:synap_filt}
We investigate a set of synaptic filters $\mathbf{\textit{h}} = \{h_{1}, h_{2}, h_{3}\}$ containing three different synaptic filters [Fig.~\ref{fig:full_system}(Middle block)]: $h_{1}$, the ideal \textit{high-pass} filter; $h_{2}$, the ideal \textit{low-pass} filter and $h_{3}$ the \textit{pulse wave} filter. The operation of filtering for the three different filters is detailed in Eq.~\eqref{eq:param_filter}.

We apply filter $h_{1}$ to the learned (unperturbed) network parameters $\theta$, resulting in perturbed network parameters $\tilde{\theta}_{1, \alpha_i}$ for a given $\alpha_{i}$ threshold, as per:
\begin{equation}
\label{eq:step_filter_func}
    \tilde{\theta}_{1,\alpha_i} = h_{1}(\theta, \alpha_i) = \left\{
    \begin{array}{rl}
         0 & \text{if} \;\; \theta \leq \alpha_i, \\
         1 & \text{otherwise} \;\\
    \end{array}
    \right.,
\end{equation}
where $\alpha_i \in \alpha$, and $\alpha =\{\alpha_{0}, \alpha_{1}, \dots,\alpha_{A}\}$ we create $|\alpha|$ thresholds between the lower and upper bounds $\alpha_{0} = \min(\theta)$ and $\alpha_{A} = \max(\theta)$. This results in a set of filtered networks $\mathcal{S}_\alpha$ with each threshold defined as $\alpha_{i} = \alpha_{i-1} + \Delta_{\alpha}$ for a step length $\Delta_{\alpha} = [{\max(\theta)-\min(\theta)}]/{A}$ {\color{blue}(i.e. $\Delta_{\alpha} = 1/{A}$ when $\alpha$ is normalised between 0 and 1)} and ${\color{blue}i=\{i \in \mathbb{N}: 0 \leq i \leq A\}}$. 

Similar to filter $h_{1}$, we apply the filter $h_{2}$ to the learned (unperturbed) network parameters $\theta$ from the opposite direction, resulting in perturbed network parameters $\tilde{\theta}_{2,\alpha_i}$ for an $\alpha_i \in \alpha$:
\begin{equation}
\label{eq:step_filter_func_2}
    \tilde{\theta}_{2,\alpha_i} = h_{2}(\theta, \alpha_i) = \left\{
    \begin{array}{rl}
         0 & \text{if} \;\; \theta \geq \alpha_{i},\\
         1 & \text{otherwise} \;\\
    \end{array}
    \right.,
\end{equation}
where
    $\alpha_{i} = \alpha_{i-1} - \Delta_{\alpha}$, 
    $\alpha_{0} = \min(-\theta), \; \text{and } \alpha_{A} = \max(-\theta).$ 

The pulse wave filter $h_{3}$, applied to $\theta$ results in equal filtered parameters $\theta_{3,\alpha_i}$ for values of $\alpha_i$ increased from $\min(\mathbf{\theta})$ to $\max(\theta)$ or decreased from $\max(\theta)$ to $\min(\theta)$. The results of filter $h_{3}$ is given by:
\begin{equation}
\label{eq:pulse_filter_func}
    \tilde{\theta}_{3,\alpha_i} = h_{3}(\theta, \alpha_i) = \left\{
    \begin{array}{rl}
         0 & \text{if} \; \alpha_{i}-\frac{\Delta_{\alpha}}{2} < \theta \leq \alpha_{i}+\frac{\Delta_{\alpha}}{2},\\
         1 & \text{otherwise} \;\\
    \end{array}
    \right.,
\end{equation}
where
    $\alpha_{i} = \alpha_{i-1} \pm \Delta_{\alpha}$, 
    $\alpha_{0} = \min(\pm\theta), \;\text{and } \alpha_{A} = \max(\pm\theta).$
%
In Eq.~\eqref{eq:pulse_filter_func}, the threshold window shifts by $\Delta_{\alpha}$ centred at threshold $\alpha_i$ with either side having a length $\frac{\Delta_{\alpha}}{2}$. 

These three filters $h_{1}, h_{2}$, and $h_{3}$ with distinct properties when applied to a DNNs with threshold $\alpha_i$ offers three sets of distinct perturbed networks $f(\tilde{\theta}_{1,\alpha_i}, x), f(\tilde{\theta}_{2,\alpha_i}, x)$, and $f(\tilde{\theta}_{3,\alpha_i}, x)$. Therefore, we require three baseline network performances corresponding to the properties of the respective synaptic filters against which the three sets of perturbed networks are compared.  

\paragraph{Baseline Network Performances}
\label{subsec:base_resp}
%

We denote $\phi_{1}, \phi_{2}$ and $\phi_{3}$ to be the number of parameters filtered out by the synaptic filters $h_{1}, h_{2}$ and $h_{3}$ corresponding to filtering threshold $\alpha_{i}$. If the synaptic filtering procedure is only applied to a local layer $l$ (e.g., only on a convolutional layer or a linear layer) then $ \phi^{(l)}$ is the maximum number of parameters in local layer $l$. For the whole network $\phi$ denote the maximum number of parameters in the network. Let us consider $\phi^{(l)}_{1}$ to the number parameters filtered out by the filter $h_{1}$ for the layer $l$ at threshold $\alpha_{i}$, then the base network performance $\bar{\phi}^{(l)}_{1,\alpha_{i}}$ at threshold $\alpha_{i}$is given as:
\begin{equation}
\label{eq:base_sys_res}
    \overline{\phi}^{(l)}_{1,\alpha_i} = 1 - \frac{\phi^{(l)}_{1,\alpha_i}}{\phi^{(l)}},
\end{equation}
Similarly, the baseline network performances for filters $h_2$ and $h_3$ are $\overline{\phi}^{(l)}_{2,\alpha_i}$ and $\overline{\phi}^{(l)}_{3,\alpha_i}$. In Eq.~\eqref{eq:base_sys_res}, the fraction $\frac{\phi^{(l)}_{1}}{\phi^{(l)}}$ is a ratio between the number of parameters removed to the total number of parameters in the layer, defining the \textit{compactness} of the filtered layer. 

\begin{figure}[h]
\centering
\begin{subfigure}{0.48\textwidth}
  \centering
  \includegraphics[width=0.98\textwidth]{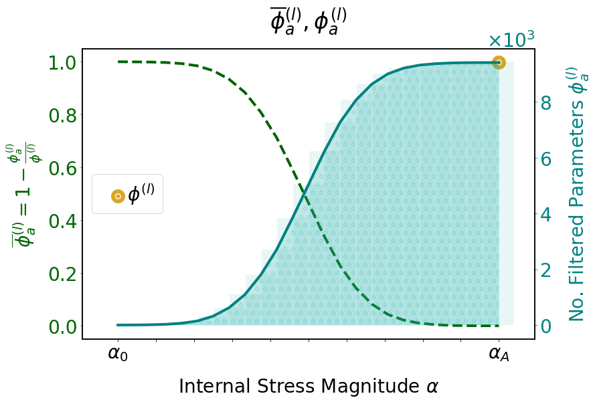}
  \caption{Baseline Network Performance.}
  \label{subfig:prop_12_1}
\end{subfigure}%
\begin{subfigure}{.47\textwidth}
  \centering
    \includegraphics[width=0.98\textwidth]{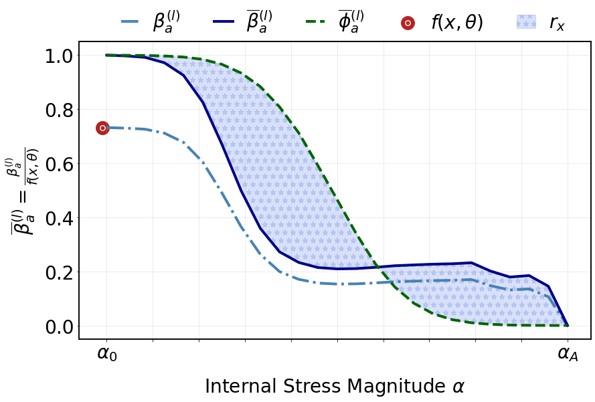}
  \caption{Baseline Network Performance and Scaled Synaptic Filtering Performance.}
  \label{subfig:prop_12_2}
\end{subfigure}\\
\caption{(a) Baseline network performance (green dotted line) $\bar{\phi}^{(l)}_{1}$ [Eq.~\eqref{eq:base_sys_res}] for ResNet-18 trained for 100 epochs on CIFAR10. $\phi^{(l)}_{1}$ is the function that contains the number of parameters filtered (teal solid line) for filtering thresholds in $\alpha$ for filter $h_{1}$ on layer $l$. The maximum number of parameters in layer $l$ is denoted by $\phi^{(l)}$ (yellow dot). (b) Comparison of the scaled of synaptic filtering performance with baseline network performance, and synaptic robustness computation. The network performance to the synaptic filter is $\beta^{(l)}_{1}$ (blue dotted line), which is scaled w.r.t the unperturbed network accuracy $f(x,\theta)$ (red dot), resulting in $\bar{\beta}^{(l)}_{1}$ (blue solid line). The blue shaded region $r_{x}$, enclosed by base system response $\bar{\phi}^{(l)}_{1}$ (green dotted line) is the area [Eq.~\eqref{eq:synp_rob_score_clean} and Eq.~\eqref{eq:synp_rob_score_adv}] of synaptic robustness.}
\label{fig:anal_proc_1_2}
\end{figure}

We consider the baseline network performance for all values of  $\alpha$, which we use to determine the parameter characteristics to synaptic filtering, as a function that reduces the network performance proportionally to the internal stress (synaptic filtering) applied on the network (see Fig.~\ref{subfig:prop_12_1}). Using this definition of the baseline network performance, we expect the network performance to decrease proportionally to the number of parameters filtered by the synaptic filtering procedure (see Fig.~\ref{subfig:prop_12_2}). The underling assumption of the baseline network performance is that the parameters being filtered out have an overall influence on the network performance. Hence, the baseline network performance represents the expected behaviour of the network, given as the classification accuracy on the test set, whilst the network is subjected the synaptic filtering procedure for all synaptic filtering threshold values $\alpha$. 


\paragraph{Network Compactness}
\label{subsubsec:network_compact}
Our synaptic filtering method is a systematic ablation of DNN parameters to analyze variations in network performance caused by parameter filtering. We show that a proportion of the network parameters can be filtered out from a DNN, whilst retaining (and occasionally improving) the network performance on both clean and adversarially perturbed test sets~\cite{Siraj2019Robust}. The characteristics of the baseline network performance, describes a network with parameters that, when filtered, proportionally influence the network performance. From Eq.~\eqref{eq:base_sys_res}, the proposed baseline network performance is linked to the the compactness of the network/layer; the characteristics of the baseline network performance is inversely proportional to the compactness ratio of the network/layer weights. For a specific non-random synaptic filtering method, the compactness characteristics of a network is constant for different variations to the input (e.g. adversarial attacks). Thus, we can compare the scaled network performances of a network to both clean and adversarial datasets, against the baseline network performance.


\paragraph{Network vs. adversary}
\label{subsec:network_vs_adversary}

For a network, we define the network performances for all synaptic filtering thresholds $\alpha$ to be an $|\alpha|$-length vector of the network prediction accuracies $p(\hat{y}_{\alpha}=y|f(x,\tilde{\theta}_\alpha))$ on the test set $x$. The network performance to the synaptic filtering $h_{1}$ [Eq.~\eqref{eq:step_filter_func}], $h_{2}$ [Eq.~\eqref{eq:step_filter_func_2}] and $h_{3}$ [Eq.~\eqref{eq:pulse_filter_func}] are given as $\beta_1$, $\beta_2$ and $\beta_3$ respectively. We construct a clean network performance matrix $\beta$ on inputs $x$ by combining $\beta_1$, $\beta_2$ and $\beta_3$ as:
\begin{equation}
\label{eq:comb_filter_res}
   \beta = \begin{bmatrix}
           \beta_{1} \\
           \beta_{2} \\
           \beta_{3}
         \end{bmatrix} = \begin{bmatrix}
           p(\hat{y}_{\alpha}=y|f(\tilde{\theta}_{1}, x)) \\
           p(\hat{y}_{\alpha}=y|f(\tilde{\theta}_{2}, x)) \\
           p(\hat{y}_{\alpha}=y|f(\tilde{\theta}_{3}, x))
         \end{bmatrix}.
\end{equation}
We further apply the synaptic filtering to the network under an adversarial attack $\delta$ with perturbation magnitudes $\epsilon$, resulting in adversarial network performance matrix $\beta_{\epsilon}$.

\begin{equation}
\label{eq:comb_filter_res_adv}
  \beta_{\epsilon} = \begin{bmatrix}
          \beta_{1,\epsilon} \\
          \beta_{2,\epsilon} \\
          \beta_{3,\epsilon}
         \end{bmatrix} = \begin{bmatrix}
          p(\hat{y}_{\alpha}=y|f(\tilde{\theta}_{1}, x_{\delta_{\epsilon}})) \\
          p(\hat{y}_{\alpha}=y|f(\tilde{\theta}_{2}, x_{\delta_{\epsilon}})) \\
          p(\hat{y}_{\alpha}=y|f(\tilde{\theta}_{3}, x_{\delta_{\epsilon}}))
         \end{bmatrix}
\end{equation}

\paragraph{Targeted parameters}
\label{subsubsec:tar_params}

\begin{figure}[!ht]
    \centering
    \includegraphics[width=0.75\textwidth]{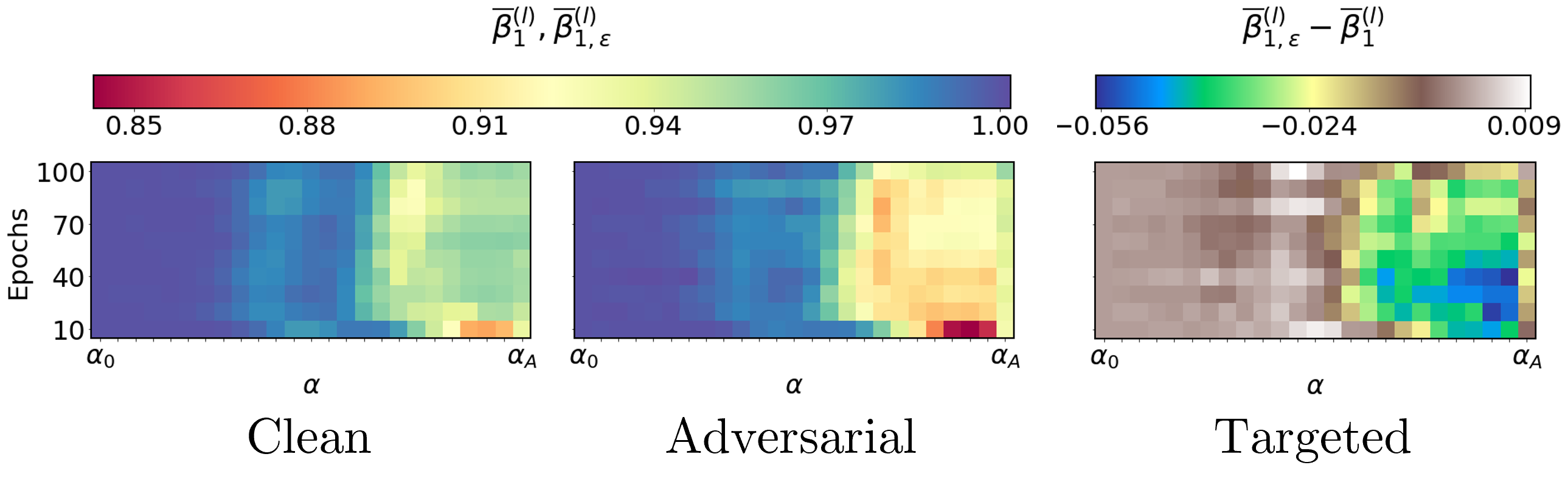}
 \caption{Learning \textit{landscape} of layers and the \textit{regime change} of test accuracy. Targeted parameters of ResNet-18 trained on MNIST using filter $h_{1}$. Showing the combined responses for layer `layer3.0.conv2', measured every 10 epoch up to 100 epochs. The difference between clean (left) and adversarial (middle) responses results in the targeted parameters (right). Every pixel on the clean and adversarial images represents the network test accuracy and for targeted image it is the difference between former two over all evaluated epochs and $\alpha$.}
    \label{fig:adv_tar}
\end{figure}

The matrices $\beta$ and $\beta_{\epsilon}$ are the network performances on clean and adversarial datasets to the synaptic filtering method that are the two different DNN states to compared. Thus, through a comparison of $\beta$ and $\beta_{\epsilon}$ (see Fig.~\ref{fig:adv_tar}), we expose the specific parameters (targeted parameters) that are either negatively, invariently or positively affecting the synaptic filtering performances for the adversarial dataset, compared to the clean dataset. 

\subsubsection{Combined network performance of synaptic filters}
\label{subsec:com_system_res}
\begin{figure}[!ht]
    \centering
    \includegraphics[width=1\textwidth]{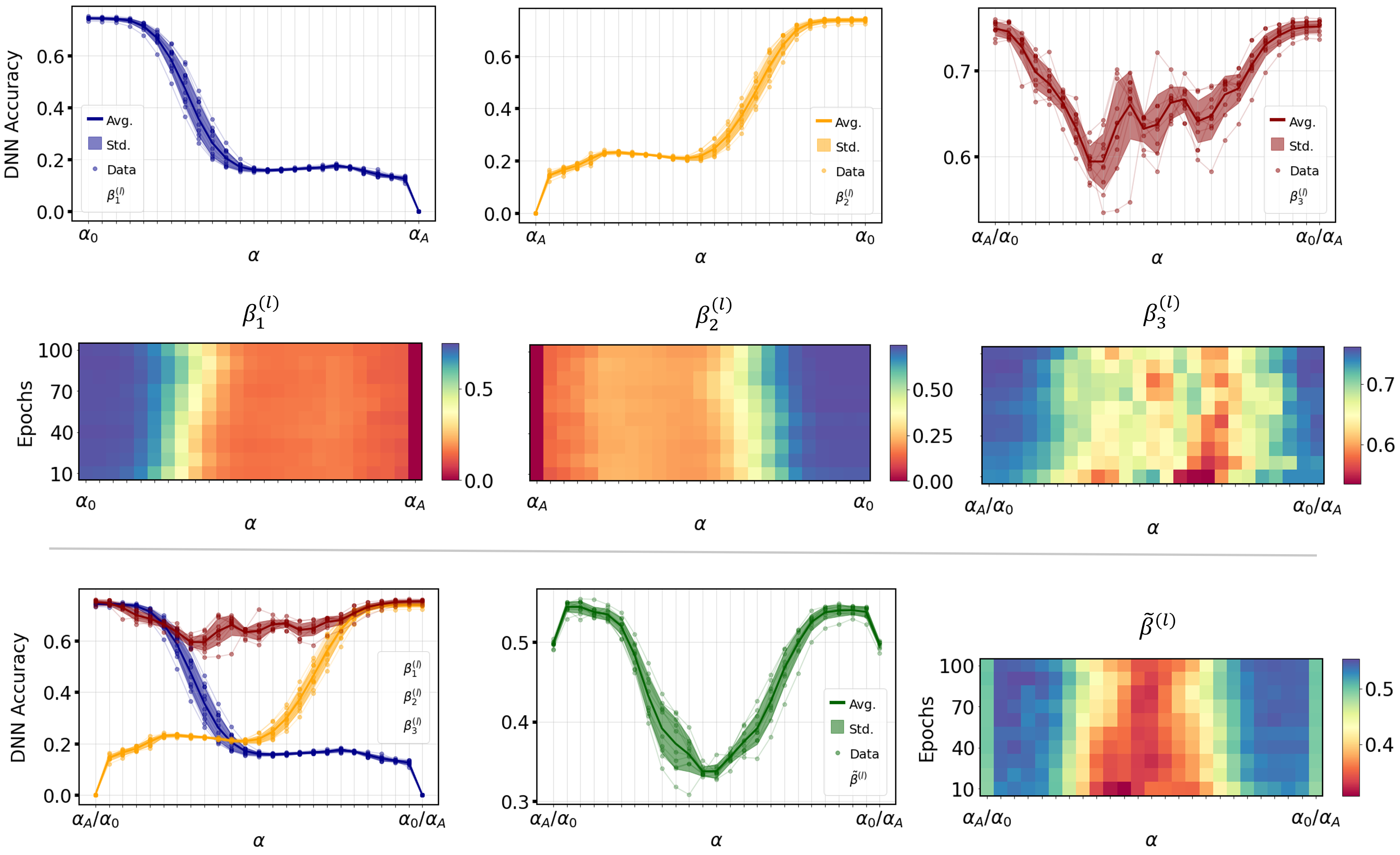}
    \caption{Example of network accuracy results of the synaptic filtering procedure applied to layer 'conv1' of ResNet-18 trained on CIFAR10, shown to illustrate the combined system response. {\color{blue}The bottom-left plot is a combination of three top-row plots.}}
    \label{fig:resnet18_comb_ex}
\end{figure}

Different synaptic filtering methods expose different characterisations of parameters of the network. Thus we combine the network performances of different synaptic filters using a function $g(\cdot)$ to form a combined network performance $\tilde{\beta}$, as shown in the synaptic filtering framework in Fig.~\ref{fig:full_system}(right). In order to combine the performances, let us consider $\beta$ as the network performance to be combined; the procedure is the same for the adversarial network performances to the synaptic filters $\beta_{\epsilon}$. We take $\bar{\beta}$ as the performance of the perturbed network (synaptic filtering performance) relative to the unperturbed network performance $f(x,\theta)$. Subsequently, we take the mean of the performances of the network of all three different filters, as such: 
%
\begin{equation}
\label{eq:comb_resp_norm}
    \tilde{\beta}_i = g(\overline{\beta}_i) = \frac{1}{|h|} \sum_{j\in h} \overline{\beta}_{j,i} \quad \text{ for } i = 1,\ldots, |\alpha|.
\end{equation}
%
Fig. \ref{fig:resnet18_comb_ex} shows an example of network accuracy results of the synaptic filtering procedure applied to layer `conv1' of ResNet-18 trained on CIFAR10. The top row shows the effects of three different filters on network accuracy; the middle row shows each filter's epoch-wise effect on network accuracy. The third row is the effect of the combined response [as per Eq.~\eqref{eq:comb_resp_norm}] of the filters.

\begin{figure}
    \centering
    \includegraphics[width=1\textwidth]{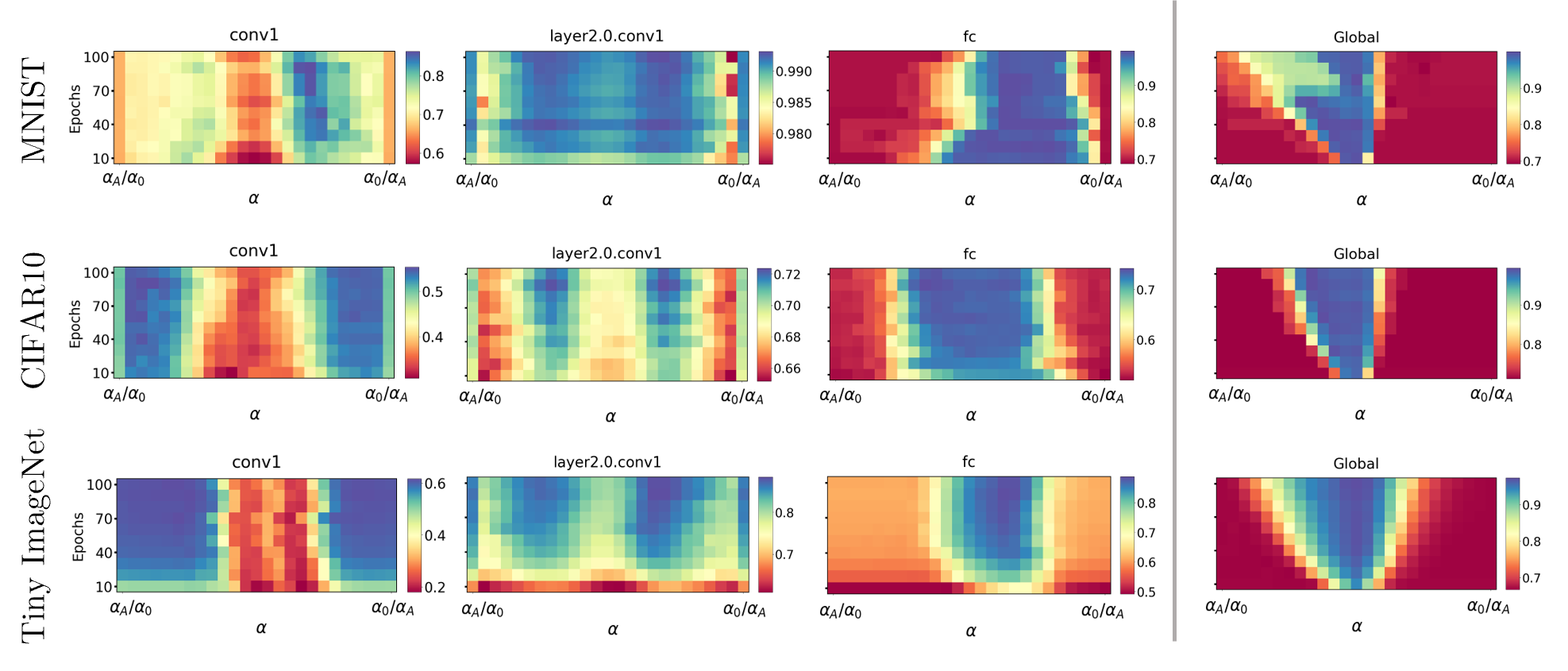}
    \caption{Combined synaptic filtering responses for ResNet-18 trained on CIFAR10, MNIST and Tiny ImageNet datasets for every 10 epochs up to 100 epochs. (\textbf{1}) Local layer-wise system response to the filtering methods for all $\alpha$ values. (\textbf{2}) Global network responses using the full network for all $\alpha$ values. Pixel intensities on the shown images represents the average network accuracy using the different synaptic filters on the clean test dataset, for each $\alpha_{i}$ in $\alpha$.}
    \label{fig:resnet18_comb}
\end{figure}

Similarly, the combined adversarial network performance $\tilde{\beta}_{\epsilon}$ is computed by replacing  $\bar{\beta}$ with $\bar{\beta}_{\epsilon}$ in Eq.~\eqref{eq:comb_resp_norm}, where $\bar{\beta}_{\epsilon}$ is the performance of the perturbed network (synaptic filtering performance) relative to the unperturbed network performance $f(x_\epsilon,\theta)$ on adversarial perturbed datasets $x_\epsilon$. Although the combined network performances $\tilde{\beta}$ and $\tilde{\beta}_{\epsilon}$ offer {\color{blue} more} descriptive information to examine the network parameters, a single synaptic filter is also able to expose the targeted network parameters. As calculating $\tilde{\beta}$ and $\tilde{\beta}_{\epsilon}$ is computationally expensive for local analysis (as this increases exponentially to the number of local layers in a DNN), we suggest computing $\tilde{\beta}$ and $\tilde{\beta}_{\epsilon}$ for all network parameters (i.e., global analysis). Fig. \ref{fig:resnet18_comb} shows the combined synaptic filtering responses (local response is in the left three columns and the global response is on the rightmost column in Fig. \ref{fig:resnet18_comb}) for ResNet-18 trained on MNIST, CIFAR10, and Tiny ImageNet datasets (see each row in Fig. \ref{fig:resnet18_comb} for respective dataset) for every 10 epochs up to 100 epochs). 

\subsection{Parameter scoring for DNN parameter characterization}
\label{subsec:pahse_2}
To expose the network parameters targeted by the adversary, let us consider the network performance $\beta^{(l)}_{1}$ for synaptic filter $h_{1}$ on layer $l$. We scale $\beta^{(l)}_{1}$ relative to $f(x,\theta)$ resulting in $\overline{\beta}^{(l)}_{1}$; the baseline network performance is $\overline{\phi}^{(l)}_{1}$ [Eq.~\eqref{eq:base_sys_res}] and the procedure is captured in Fig.~\ref{fig:anal_proc_1_2}. Similarly, we compute $\overline{\beta}_{2}^{(l)}$ and $\overline{\beta}_{3}^{(l)}$ for synaptic filters $h_{2}$ and $h_{3}$. The combined performance of the three different synaptic filters is $\hat{\bar{\beta}}^{(l)}$ [Eq.~\eqref{eq:comb_resp_norm}]. 
\begin{figure}[!b]
\centering
\begin{subfigure}{0.475\textwidth}
  \centering
  \includegraphics[width=0.97\textwidth]{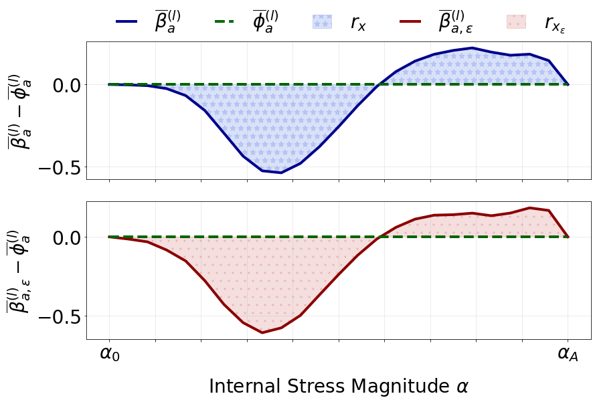}
  \caption{Clean and Adversarial Performances}
  \label{subfig:prop_34_3}
\end{subfigure}%
\begin{subfigure}{.48\textwidth}
  \centering
    \includegraphics[width=0.98\textwidth]{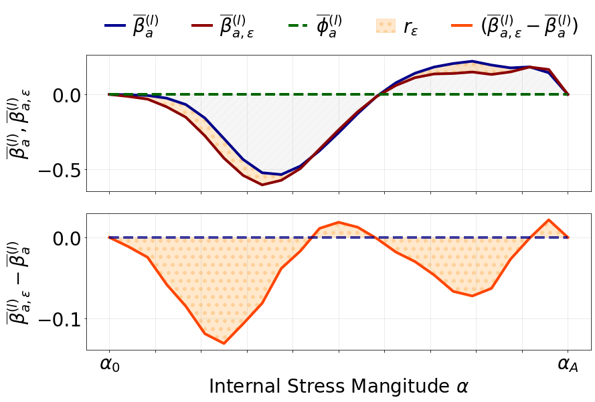}
  \caption{Parameter Score Computation}
  \label{subfig:prop_34_4}
\end{subfigure}\\
\caption{(a) Synaptic parameter score computation for clean and adversarial inputs (ResNet-18, CIFAR10, 100 epochs, layer 'conv1'). The scaled network performance to the clean $\bar{\beta}^{(l)}_{1}$ (top) and adversarial $\bar{\beta}^{(l)}_{1,\epsilon}$ (bottom) datasets. The clean and adversarial parameter scores are $r_{x}$ and $r_{x_{\epsilon}}$. (b) The behaviour of the network responses (ResNet-18, CIFAR10, 100 epochs, layer 'conv1') using synaptic filtering on the CIFAR10 clean $\bar{\beta}^{(l)}_{1}$ and adversarial $\bar{\beta}^{(l)}_{1,\epsilon}$ datasets over normalised $\alpha$ (top). The area $r_{\epsilon}$ [Eq.~\eqref{eq:r_epsilon}] is the adversarial parameter score (bottom).}
\label{fig:anal_proc_3_4}
\end{figure}

\subsubsection{Parameter score for clean data}
\label{subsubsec:synap_rob}

We take the baseline network performance $\bar{\phi}^{(l)}_{1}$ [Eq.~\eqref{eq:base_sys_res}] for synaptic filter $h_{1}$ as the point to which we evaluate the filtered network responses to. We take $\bar{\phi}^{(l)}_{1}$ to describe a network/layer that contains neither an excess nor a deficiency of parameters that influence the network performance (i.e. the removal of any parameter affects the network performance). The network performance, on average will react \textit{inversely proportional} to ablation of network parameters to synaptic filtering. The parameter score to synaptic filtering for a network using a clean dataset is $r_{x}$ is shown in Fig.~\ref{subfig:prop_34_3}(top)] and given as:
\begin{equation}
\label{eq:synp_rob_score_clean}
    r_{x} = \sum_{i=0}^{{\color{blue}A}} (\bar{\beta}^{(l)}_{1} - \bar{\phi}^{(l)}_{1}) \Delta_{\alpha},
\end{equation}
Where $\Delta_{\alpha}$ is the change in the $\alpha$ threshold window. A parameter score \textit{equal to} 0 signifies that the network/layer responds, on average, \textit{proportionally} to synaptic filtering, i.e., proportional to variations in architecture and thus is considered \textit{robust}. Where the score $r_{x}$ is \textit{less than} 0, this indicates that the network/layer contains \textit{fragile} parameters to the network performance. Conversely, where the value of $r_{x}$ is \textit{greater than} $0$, the parameter score indicates that the network/layer contains \textit{antifragile} parameters, where the removal of parameters from the network/layer results in a network performance that is better than the baseline network performance.

\subsubsection{Parameter score for adversarial data}
\label{subsubsec:synap_adv}
The parameter score to synaptic filtering for a network using an adversarial dataset is $r_{x_{\epsilon}}$, and is calculated using the baseline network performance $\bar{\phi}^{(l)}_{1}$. The baseline network performance is compared with the adversarial dataset performance $\bar{\beta}^{(l)}_{1,\epsilon}$ to give the parameter characterization score $r_{x_{\epsilon}}$, as per:
\begin{equation}
\label{eq:synp_rob_score_adv}
    r_{x_{\epsilon}} = \sum_{i=0}^{{\color{blue}A}} (\bar{\beta}^{(l)}_{1,\epsilon} - \bar{\phi}^{(l)}_{1}) \Delta_{\alpha},
\end{equation}
Where $\Delta_{\alpha}$ is the change in the $\alpha$ threshold window. A parameter score \textit{equal to} 0 signifies that the network/layer responds, over all magnitudes of internal stress, \textit{proportionally} to synaptic filtering, i.e., proportional to variations in architecture and thus is considered \textit{robust}. Where scores $r_{x}$ and $r_{x_{\epsilon}}$ are \textit{less than} $0$, this indicates that the network/layer contains \textit{fragile} parameters w.r.t. the network performance. Conversely, where $r_{x}$ and $r_{x_{\epsilon}}$ are \textit{greater than} $0$, the scores indicate that the network/layer contains \textit{antifragile} parameters.

\subsubsection{Difference of parameter scores}
\label{subsubsec:adv_rob}
To compute the effects of the adversarial attack on the parameter characterisation, using our proposed synaptic filtering method, we take the baseline network performance to be the synaptic filtering performance on the clean dataset ($\bar{\phi}^{(l)}_{1} = \bar{\beta}^{(l)}_{1}$). The difference in the adversarial dataset performance $\overline{\beta}^{(l)}_{1, \epsilon}$ and clean dataset performance $\overline{\beta}^{(l)}_{1}$ (baseline network performance), results in the effects of the adversary on the synaptic filtering performance of the network. We take the \textit{area of the residual} as the effects of the adversary on the network. The value of $r_{\epsilon}$ is computed by taking the discrete area difference, as shown in Fig.~\ref{subfig:prop_34_4} (bottom) and expressed as:
\begin{equation}
\label{eq:r_epsilon}
    r_{\epsilon} = \sum_{i=0}^{{\color{blue}A}} (\overline{\beta}^{(l)}_{1,\epsilon} - \overline{\beta}^{(l)}_{1}) \Delta_{\alpha}.
\end{equation}
If the network performs equally to clean and adversarial datasets for all filtering thresholds $\alpha$, the value of $r_{\epsilon} = 0$. Where $r_{\epsilon} < 0$, the network performance on the adversarial dataset is greater than the network performance on the clean dataset. This signifies that the evaluated network/layer contains parameters that increase the network performance on the adversarial dataset compared to the clean dataset. Conversely, $r_{\epsilon} > 0$ signifies that the network performance on the clean dataset is greater than the network performance on the adversarial dataset. This signifies that the evaluated network/layer contains parameters that decrease the network performance on the adversarial dataset compared to the clean dataset. Hence, the magnitude of $r_{\epsilon}$ gives us a scalar value of the difference in clean and adversarial responses to the filtering.

\subsection{Experimental set-up}
\label{subsec:exp_setup}
Our experiment setting includes standard training of state-of-the-art DNNs on popular benchmark datasets. 
\paragraph{State-of-the-art datasets used}
All experiments\footnote{Source code: https://github.com/SynapFilter/InferLink} in this study are performed on the MNIST~\cite{lecun-mnisthandwrittendigit-2010}, CIFAR10~\cite{krizhevsky2009learning} and Tiny ImageNet~\cite{le2015tiny} datasets. The MNIST and CIFAR10 datasets both respectively contain 80,000 examples in the training set and 10,000 examples in the test set. The Tiny ImageNet dataset contains 80,000 training and 20,000 test examples from the original training set~\cite{le2015tiny}. 

\paragraph{State-of-the-art DNNs studies}
On the benchmark datasets, we train ResNet-18, ResNet-50~\cite{He_2016_CVPR}, SqueezeNet v1.1~\cite{squeezenet2016iandola} and ShuffleNet V2 x1.0~\cite{Ma_2018_ECCV}. Each network was trained for 100 epochs, and the model of every 10 epochs was stored for analysis of our methodology. We investigated all convolutional and fully connected layers of ResNet-18, ResNet-50, SqueezeNet v1.1 and ShuffleNet V2 x1.0 only, any intermediary functions, such as the batch normalization layers, activation functions and pooling layers remain unaltered. 


\paragraph{Training of DNNs on clean datasets}
For the training, the parameters of each DNN were initialised using the Kamming Uniform~\cite{He_2015_ICCV} method. We use a cross-entropy loss function and the Adam optimizer~\cite{kingma2015adam} configured with $\gamma=0.001$, $\beta_{1} = 0.9$, $\beta_{2} = 0.999$, $\lambda = 0$ and $\varepsilon = 1\times10^{-08}$ for training networks. We train the networks using clean datasets only and apply the adversarial attack only to the test datasets for analysis using the synaptic filtering methodology. Networks are saved at every 10 epochs during network training to a maximum of 100 epochs. Saved networks are subsequently evaluated using the proposed synaptic filtering methodology, the results presented in Sec.~\ref{sec:res_anal} are shown for the saved networks.  


\paragraph{Adversarial attack on datastes}
For the adversarial attack, we use the single-step FGSM attack~\cite{goodfellow2015explaining} and analyze the difference in network performance on the test set to the proposed synaptic filtering methods (Sec.~\ref{sec:synaptic_filtering}). The effectiveness of an adversarial attack on a given dataset can be attributed  to the complexity of the datasets the attack has been applied to.

\paragraph{Collection of results}
We normalize the $r_{x}$ and $r_{x_{\epsilon}}$ parameter score values from Sec.~\ref{subsubsec:synap_rob} to be between -0.5 (indicating fragility) and 0.5 (indicating antifragility) with the mid-point being 0 (indicating robustness). We carry out the same normalization procedure independently for all $r_{\epsilon}$ values from Sec.~\ref{subsubsec:adv_rob} to be between -0.5 and 0.5. 

For each network and dataset, the synaptic filtering responses are averaged over three different randomly initialised (as per~\cite{he2015delving}) and trained networks.  In order to satisfy constraint 3 from Sec.~\ref{sec:synaptic_filtering}, we use a line search algorithm to find the optimal $\epsilon$ value for each model and dataset, that satisfies: $f(x+\delta_{\epsilon},\mathbf{W}) \approx 0.5 \cdot f(x,\mathbf{W})$. {\color{blue}When carrying out the synaptic filtering procedure, we select $A = 25$ for all experiments carried out. Therefore, the filtering step size $\Delta_\alpha = 0.04$ over the normalised range of parameters in the evaluated network/layer. Where computational resources permit, we recommend using larger values of $A$ for experimentation in order to more accurately compute the parameter scores.}

\section{Results and Analysis}
\label{sec:res_anal}
\begin{figure}[ht]
\centering
\begin{subfigure}{0.45\textwidth}
  \centering
  \includegraphics[width=0.93\textwidth]{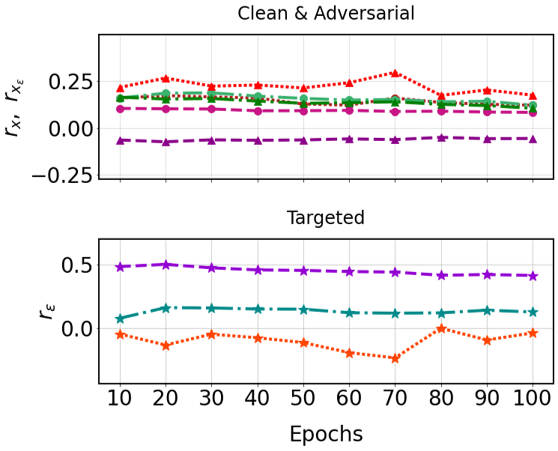}
  \caption{ResNet-18}
  \label{fig:global_resnet18}
\end{subfigure}%
\begin{subfigure}{.45\textwidth}
  \centering
    \includegraphics[width=0.93\textwidth]{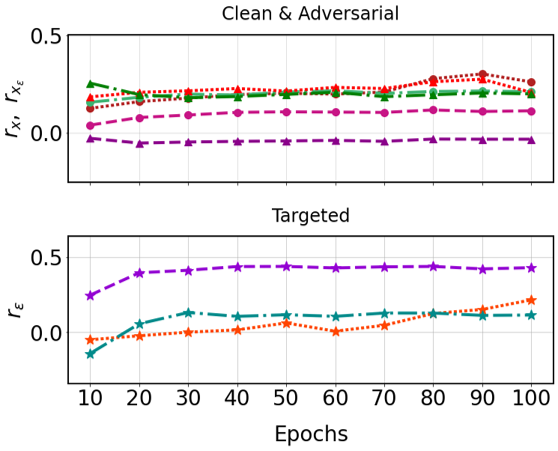}
  \caption{ResNet-50}
  \label{fig:global_resnet50}
\end{subfigure}
\hfill

\begin{subfigure}{.45\textwidth}
  \centering
  \includegraphics[width=0.93\textwidth]{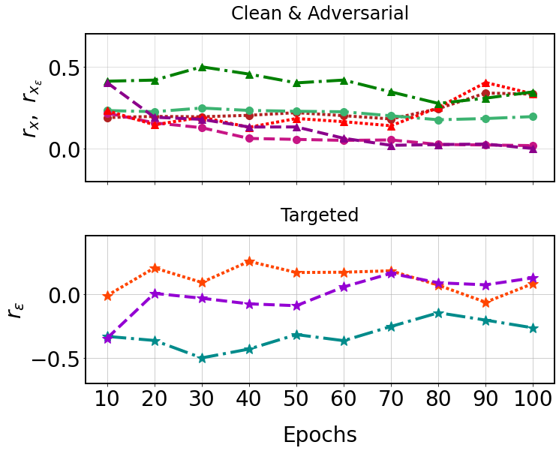}
  \caption{SqueezeNet-v1.1}
  \label{subfig:global_squeezenet}
\end{subfigure}%
\begin{subfigure}{0.45\textwidth}
  \centering
  \includegraphics[width=0.93\textwidth]{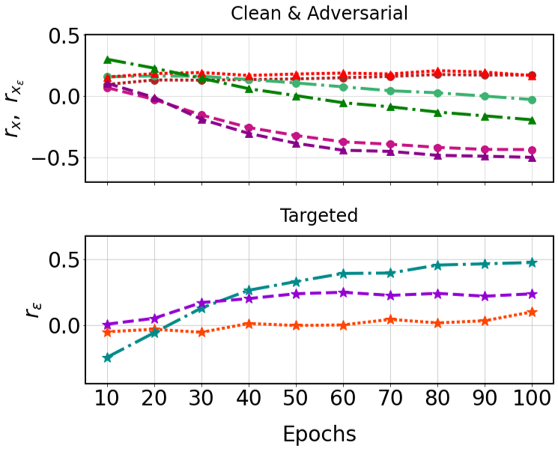}
  \caption{ShuffleNet V2 x1.0}
  \label{subfig:global_shufflenet}
\end{subfigure}


\begin{subfigure}{0.55\textwidth}
    \centering
    \includegraphics[width=0.99\textwidth]{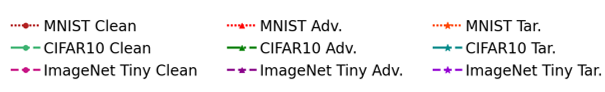}
\end{subfigure}

\caption{Global parameter scores of (a) ResNet-18, (b) ResNet-50, (c) SqueezeNet-v1.1 and (d) ShuffleNet V2x1.0 over all datasets are $r_{x}, r_{x_{\epsilon}}$ and $r_{\epsilon}$, measured every 10 epochs up to 100 epochs for the whole network parameters using synaptic filter $h_1$. The parameter score interpretation is given in Sec.~\ref{subsubsec:synap_rob} and Sec.~\ref{subsubsec:adv_rob}.}
\label{fig:rob_global_results}
\end{figure}

\begin{figure}
\centering
\begin{subfigure}{0.54\textwidth}
  \centering
  \includegraphics[width=1\textwidth]{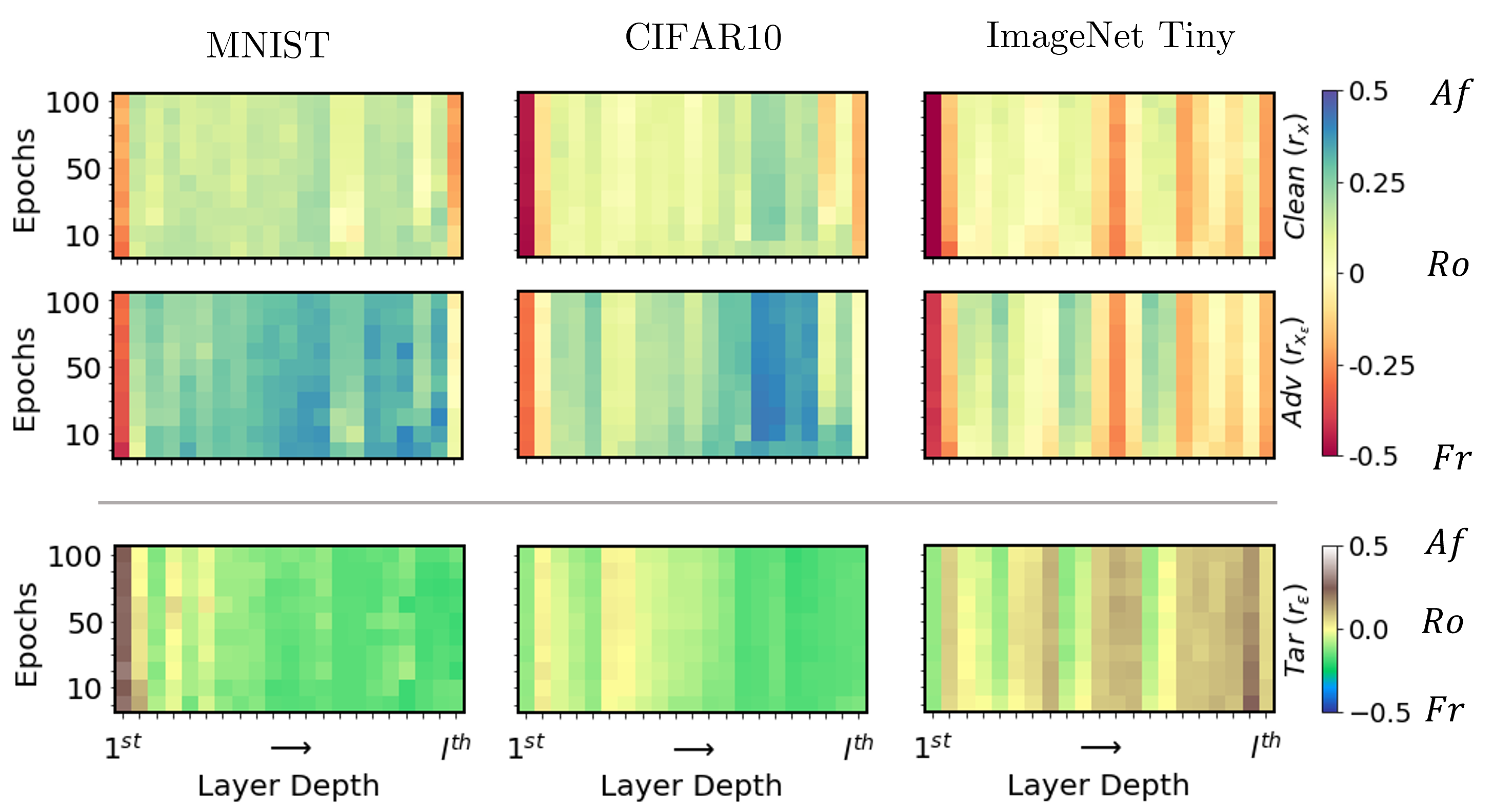}
  \caption{ResNet-18}
  \label{fig:local_sub1}
\end{subfigure}
\\
\begin{subfigure}{1\textwidth}
\centering
\includegraphics[width=1\textwidth]{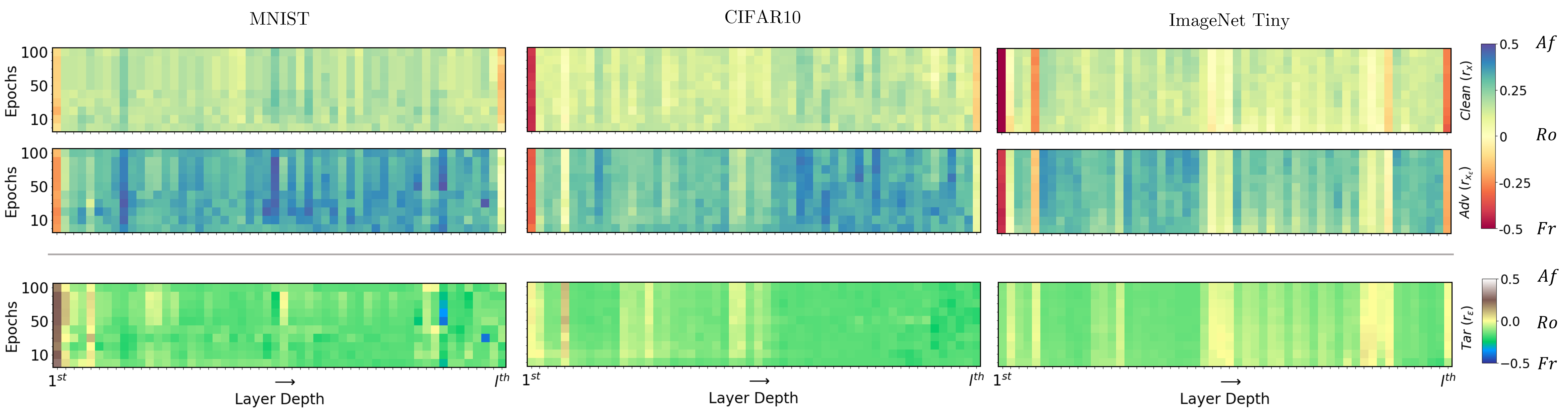}
\caption{ResNet-50}
\label{fig:local_sub2}
\end{subfigure}
\\
\begin{subfigure}{0.57\textwidth}
  \centering
  \includegraphics[width=1\textwidth]{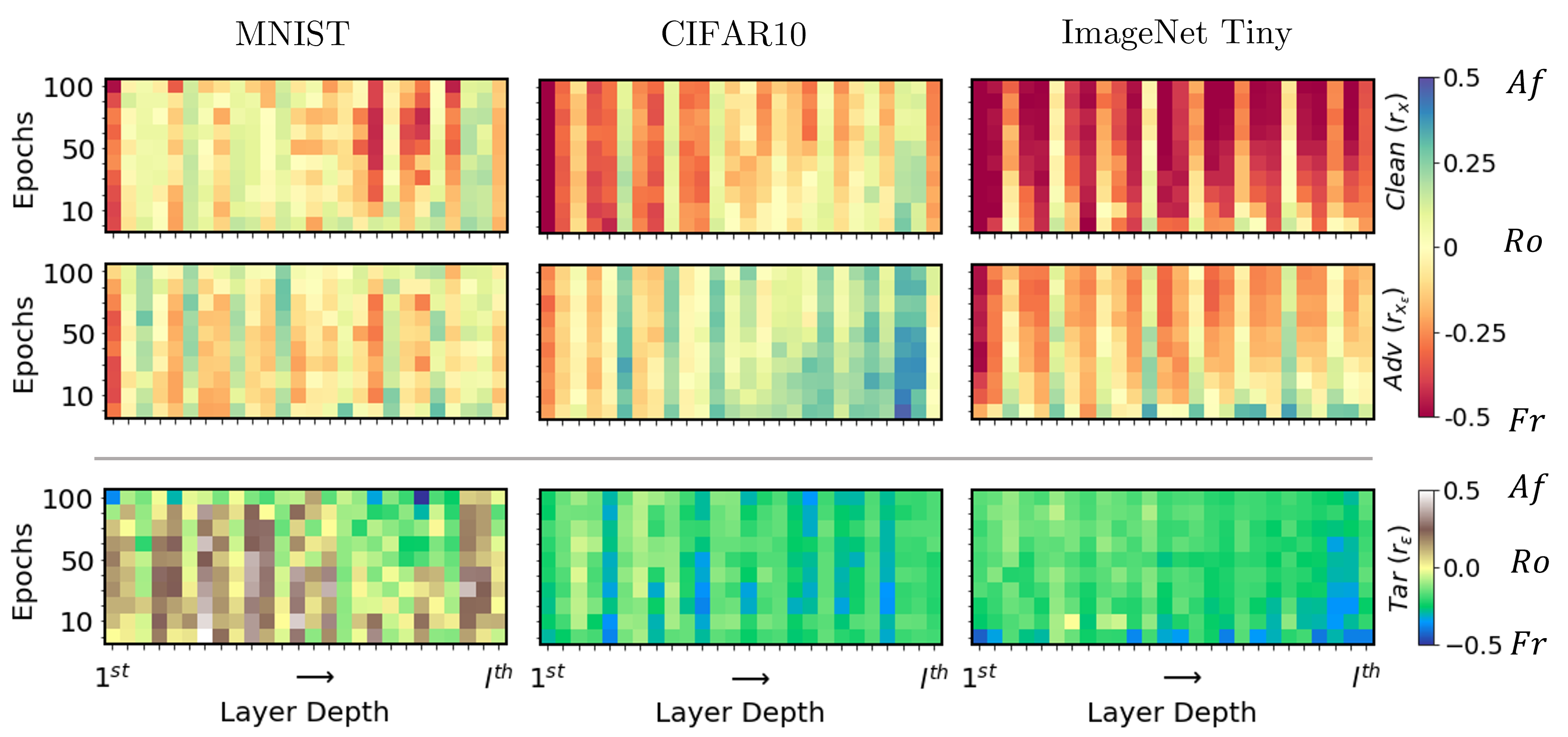}
  \caption{SqueezeNet-v1.1}
  \label{fig:local_sub3}
\end{subfigure}
\\

\begin{subfigure}{1\textwidth}
\centering
\includegraphics[width=1\textwidth]{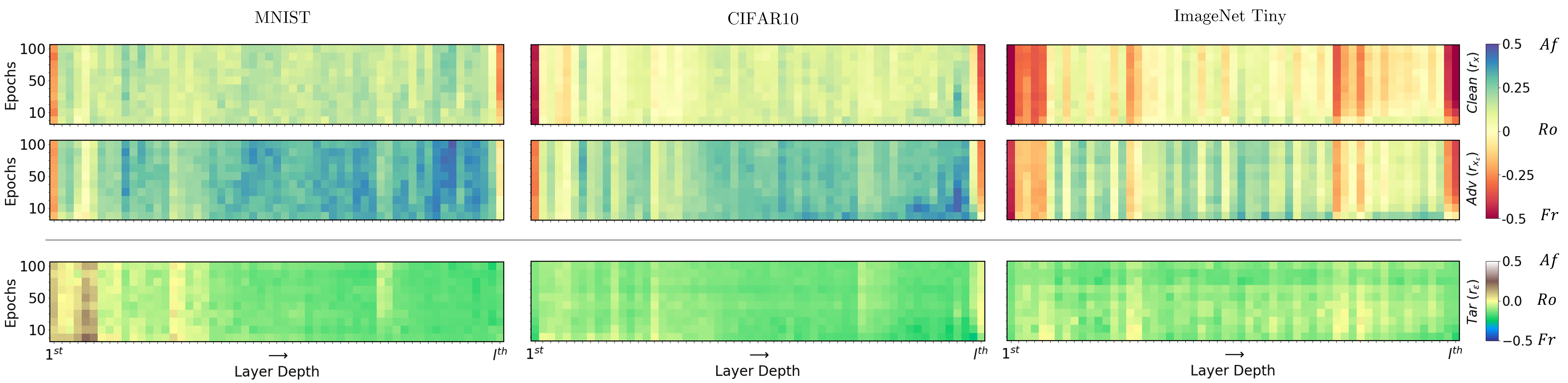}
\caption{ShuffleNet V2 x1.0}
\label{fig:local_sub4}
\end{subfigure}%

\caption{Local parameter scores of (a) ResNet-18, (b) ResNet-50, (c) SqueezeNet-v1.1 and (d) ShuffleNet V2 x1.0 over all datasets. The parameter scores $r_{x}, r_{x_{\epsilon}}$ and $r_{\epsilon}$ are measured every 10 epochs up to 100 epochs and for all layers in the network for filter $h_a$. The parameter score interpretation is given in Sec.~\ref{subsubsec:synap_rob} and Sec.~\ref{subsubsec:adv_rob}. The fragile, robust, and antifragile parameters of the network are respectively represented by values 'Fr,' 'Ro,' and 'Af' (see rightmost colour bar).}
\label{fig:resnet18_rob}
\end{figure}

The results of global (full network parameters) and local (network layer parameters) analysis shown in Figs.~\ref{fig:rob_global_results} and \ref{fig:resnet18_rob} describe the fragility, robustness, and antifragility characteristics of parameters (cf. Sec.~\ref{subsubsec:synap_rob} and Figs.~\ref{fig:resnet18_rob} and \ref{fig:rob_global_results}). Furthermore, the results show the adversarially targeted ($r_\epsilon$) parameters (cf. Sec.~\ref{subsubsec:adv_rob} and Figs.~\ref{fig:resnet18_rob} and \ref{fig:rob_global_results}). We identify parameter characteristics that are invariant for clean and adversarial datasets across different datasets and networks.  
%

\paragraph{Fragility, robustness, and antifragility} 
The global parameter scores for networks on different datasets are shown in Fig.~\ref{fig:rob_global_results}. We find that ResNet18 and ResNet50 networks exhibit invariant parameter characteristics to different datasets: particularly for $r_{x}$ and $r_{x_{\epsilon}}$ values related to the CIFAR10 and ImageNet Tiny performances, over 100 epochs. The adversarial targeting results ($r_{\epsilon}$ values) are comparable for the CIFAR10 and ImageNet Tiny responses, with $r_{\epsilon}$ values for MNIST, suggesting that the clean dataset response is consistently greater than the adversarial dataset performance. From the ShuffleNet V2x1.0 parameter scores, we find distinctions in $r_{x}$ and $r_{x_{\epsilon}}$, for the MNSIT dataset, over 100 epochs. We  see the ShuffleNet V2x1.0 parameters as transitioning from fragile to antifragile for the ImageNet Tiny dataset. From the SqueezeNet-v1.1 results for the ImageNet Tiny dataset, we observe a convergence of $r_{x}$ and $r_{x_{\epsilon}}$ to 0 over 100 epochs, indicating the network performance as robust, for both clean and adversarial datasets. 
%

The local parameter scores provide a \textit{learning landscape} to examine individual network parameters (Fig.~\ref{fig:resnet18_rob}). All of the evaluated network and dataset parameter scores exhibit \textit{invariant} fragility characteristics (marked as `Fr') at the 1-st convolutional layer and the l-th linear layer, for both clean ($r_{x}$) and adversarial ($r_{x_{\epsilon}}$) parameter scores. This is further shown in Fig.~\ref{fig:local_sub1} ImageNet Tiny; Fig.~\ref{fig:local_sub3} MNIST, CIFAR10 and ImageNet Tiny, and Fig.~\ref{fig:local_sub4} ImageNet Tiny. We see the presence of robust parameters (marked as `Ro') in Fig.~\ref{fig:local_sub1} CIFAR10 and ImageNet Tiny; Fig.~\ref{fig:local_sub2} ImageNet Tiny; Fig.~\ref{fig:local_sub3} MNIST and CIFAR10, and Fig.~\ref{fig:local_sub4} CIFAR10 ImageNet Tiny. Antifragile parameters (marked as `Af') are distinctly visible in Fig.~\ref{fig:local_sub1} MNIST and CIFAR10; Fig.~\ref{fig:local_sub2} MNIST, CIFAR10 and ImageNet Tiny; Fig.~\ref{fig:local_sub4} MNIST and CIFAR10. Furthermore, periodic robustness characteristics are shown in Fig.~\ref{fig:local_sub1} ImageNet Tiny; Fig.~\ref{fig:local_sub2} MNIST, CIFAR10 and ImageNet Tiny; Fig.~\ref{fig:local_sub3} MNIST, CIFAR10, ImageNet Tiny, and Fig.~\ref{fig:local_sub4} CIFAR10 and ImageNet Tiny.

%

\paragraph{Adversarially targeted parameters}  In Fig.~\ref{fig:adv_tar}, we present targeted parameters to an adversarial attack using the combined network response for ResNet-18 trained on MNIST. We further see targeted parameters using the parameter scores $r_{\epsilon}$ (Sec.~\ref{subsubsec:adv_rob}) from Fig.~\ref{fig:rob_global_results} and Fig.~\ref{fig:resnet18_rob}. In Fig.~\ref{fig:resnet18_rob}, we show that the network response is greater for the adversarial dataset than the clean dataset (marked by `$r_{x_{\epsilon}}$'), as shown in layers of Fig.~\ref{fig:local_sub3} CIFAR10 and ImageNet Tiny. We find instances where both the adversarial performance and clean performance are equal, indicating that the layer response is robust (marked by `Ro') and shown in Fig.~\ref{fig:local_sub1} MNIST, CIFAR10 and ImageNet Tiny; Fig.~\ref{fig:local_sub2} MNIST, CIFAR10 and ImageNet Tiny; Fig.~\ref{fig:local_sub3} MNIST, and Fig.~\ref{fig:local_sub4} CIFAR10 and ImageNet Tiny. Furthermore, we see instances of the network performances for the clean dataset being greater than that of the adversarial dataset (marked by `$r_{x}$'), shown in Fig.~\ref{fig:local_sub1} MNIST and CIFAR10; Fig.~\ref{fig:local_sub2} MNIST, CIFAR10 and ImageNet Tiny, and Fig.~\ref{fig:local_sub4} MNIST and CIFAR10. 

\paragraph{Effects of batch normalization}
\label{subsubsec:feature_collapse}
\begin{figure}[ht]
    \centering
    \begin{subfigure}{1\textwidth}
    \centering
    \includegraphics[width=1\textwidth]{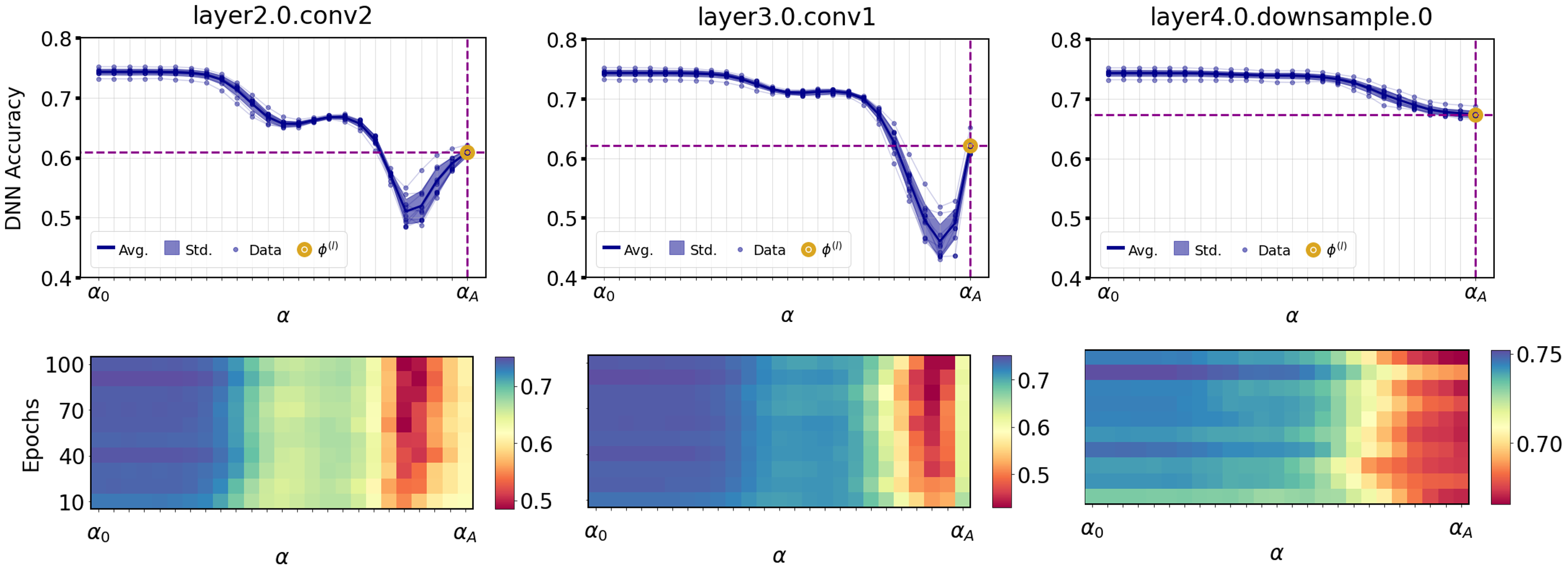}
    \caption{Filter $h_{1}$}
    \label{fig:bn_ex_1}
    \end{subfigure}\\

    \begin{subfigure}{1\textwidth}
      \centering
      \includegraphics[width=1\textwidth]{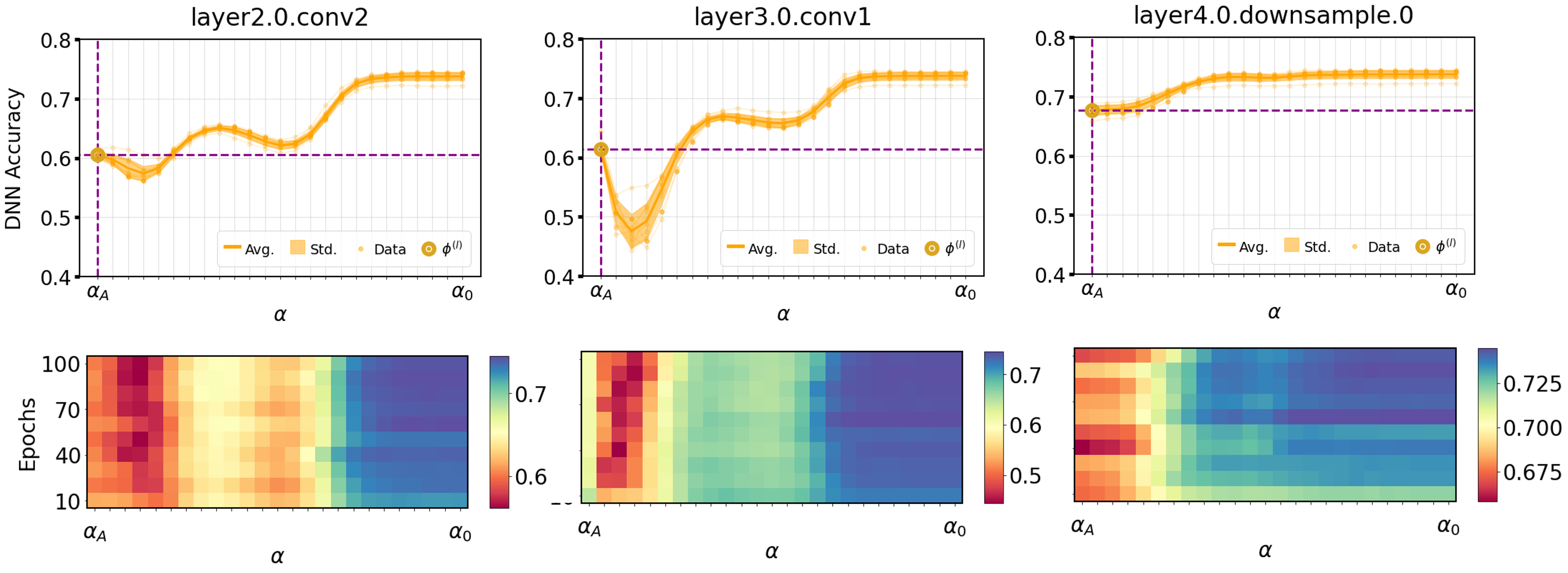}
      \caption{Filter $h_{2}$}
      \label{fig:bn_ex_2}
    \end{subfigure}\\
    \caption{Synaptic filtering network performances of ResNet-18 trained on CIFAR10 for layers 'layer2.0.conv1', 'layer3.0.conv1' and 'layer4.0.downsample.0'. The results show that, even after filtering all parameters in a layer, the network performs relatively well (shown by the purple dotted lines at $\Phi^{(l)}$, the maximum number of parameters filtered). This is due to the following batch normalization layer features propagating through the network during the forward pass, thus highlighting the effects of batch normalisation layers on network performance.}
    \label{fig:batch_norm_fig}
\end{figure}
We investigated the phenomenon of the network retaining classification performance despite all features at layer $l$ removed (see column $\alpha_A$ in Fig.~\ref{fig:adv_tar}). 
When we investigate the output of layers deeper than $l$, we discover that residual features continue to propagate through the network despite the filtering out of network weights at the $l$-th layer. This is attributed to the Batch normalization (BN) layers that follow convolutional layers and are tasked with minimising covariance shift in the network~\cite{sergey2015Batch}. When implementing a network architecture, we utilise the standard models in accordance with literature; the functionality of batch normalization layers is also predefined and remains unaltered in our analysis. Consider the condition where a convolutional layer $l$ has been filtered maximally using a synaptic filter, the subsequent batch normalization computation is given as: 
\begin{equation}
\label{eq:batch_norm}
    \hat{y}^{(l)} =
    \frac{x^{(l)}-\mathop{\mathbb{E}}[x]}{\sqrt{\mathrm{Var}[x] + \epsilon}}*\gamma^{(l)} + \beta^{(l)}
\end{equation}
Where $\hat{y}^{(l)}$ is the output of the batch normalization process at the output of convolutional layer $l$; $y^{(l-1)}$ is the output of the previous convolutional layer $l-1$ given by $f(\tilde{\theta}^{(l)}_{\{1,2,3\}}, \hat{y}^{(l-1)})$. The variables $\gamma^{(l)}$ and $\beta^{(l)}$ are learnable parameter vectors and $\epsilon$ is a value added to the denominator for numerical stability (set to $\num{1e-05}$). 
Implementations of networks compute the expectation and variance from Eq.~\eqref{eq:batch_norm} as running statistics during network training; the statistics calculated during training are used during network inference. In consequence, when the input to the BN layer following convolutional layer $l$ is a 0 vector, the case where layer $l$ has been filtered maximally through synaptic filtering, the BN layer retains features of the training batches, even when evaluating test sets. This is shown from the results in Fig.~\ref{fig:batch_norm_fig}, where the filtering of parameters from certain layers results in only a slight decrease of network performance. The ability of the network to retain sufficient performance, despite the filtering out of certain layer parameters, is due to the features propagated during the forward pass by the batch normalization layer following the filtered layer.

\paragraph{Selective backpropagation on robust and antifragile parameters}
\label{subsubsec:selective_backprop}

Upon identifying robust, fragile and antifragile parameters using the difference in parameter scores (see Sec.~\ref{subsubsec:adv_rob}) we consider fragile parameters to be parameters that, when perturbed, result in greater degradation of synaptic filtering performance on the clean dataset compared to the adversarial dataset. Robust parameters show to be invariant to both clean and adversarial datasets, and antifragile parameters show to have an increased network performance on the clean dataset compared to the adversarial dataset.

Thus, we consider fragile parameters to be parameters that are important to the network performance on the adversarial dataset. We propose selectively retraining only the robust and antifragile parameters using backpropagation. In order to carry out this operation during network training,  we take a layer-wise approach that considers the parameter characterization scores of individual network layers and we subsequently omit the characterized fragile layers corresponding to negative parameter characterizations scores from network training by zeroing out the update gradients during training. 

\begin{figure}[h!]
\centering
\begin{subfigure}{0.32\textwidth}
  \centering
  \includegraphics[width=0.98\textwidth]{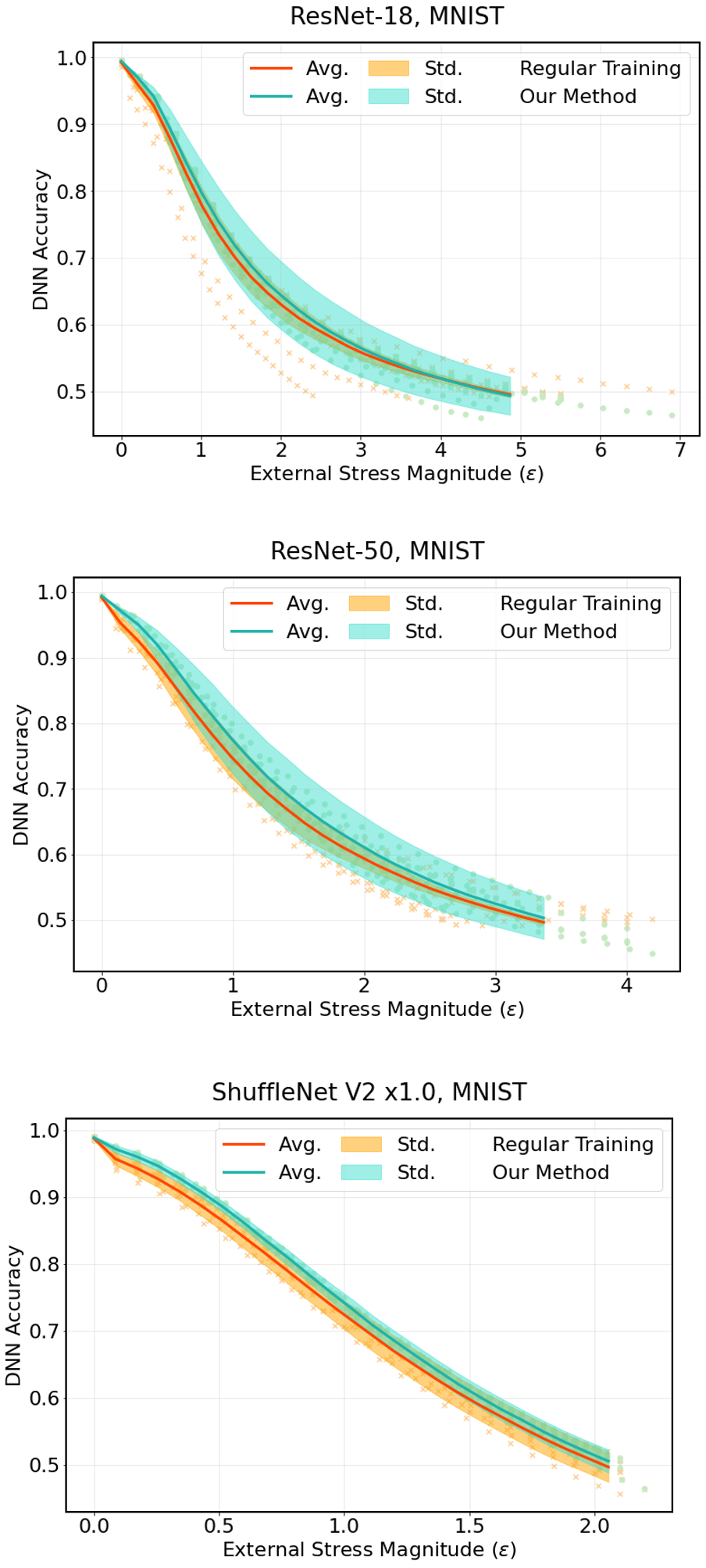}
  \caption{MNIST}
  \label{subfig:rob_increase_mnist}
\end{subfigure}%
\begin{subfigure}{.32\textwidth}
  \centering
    \includegraphics[width=0.98\textwidth]{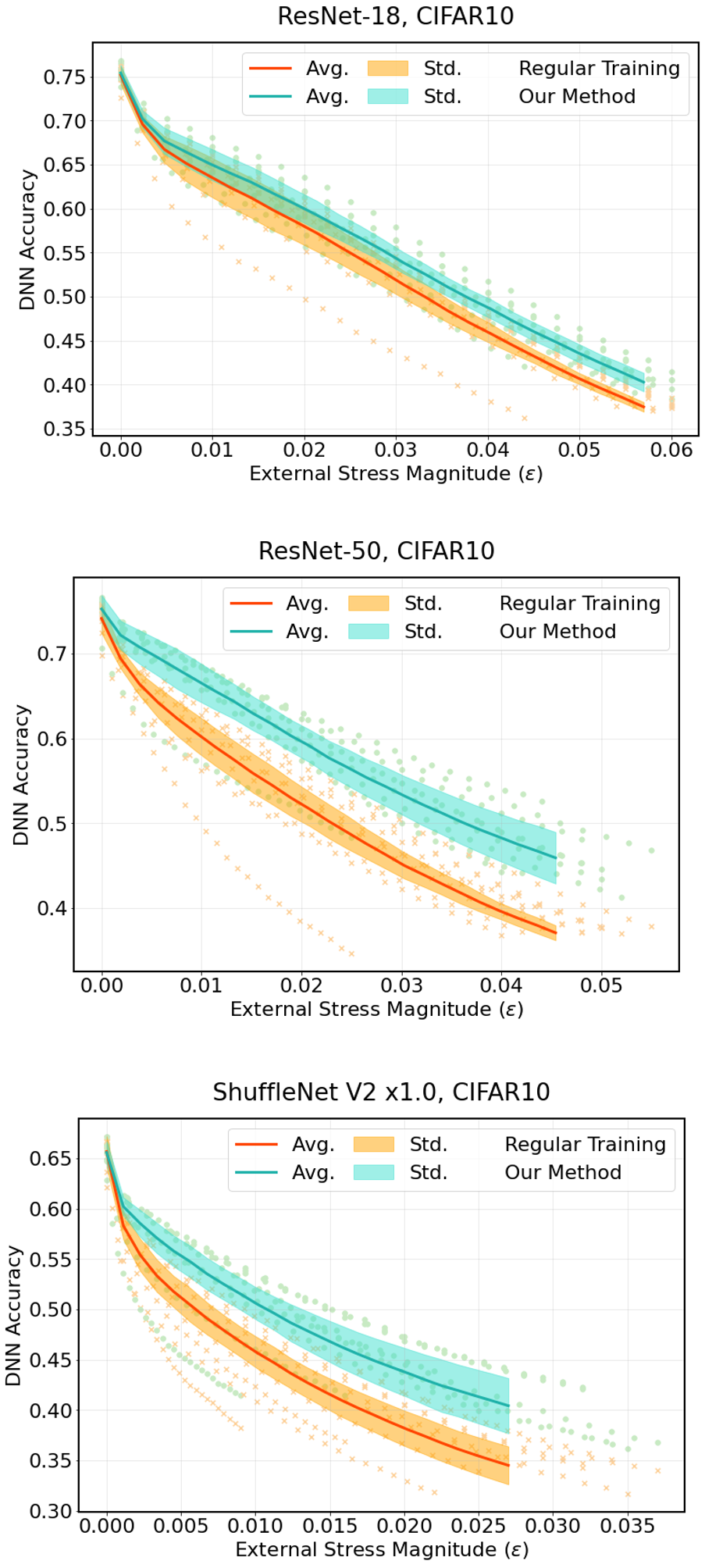}
  \caption{CIFAR10}
  \label{subfig:rob_increase_cifar10}
\end{subfigure}
\begin{subfigure}{.32\textwidth}
  \centering
    \includegraphics[width=0.98\textwidth]{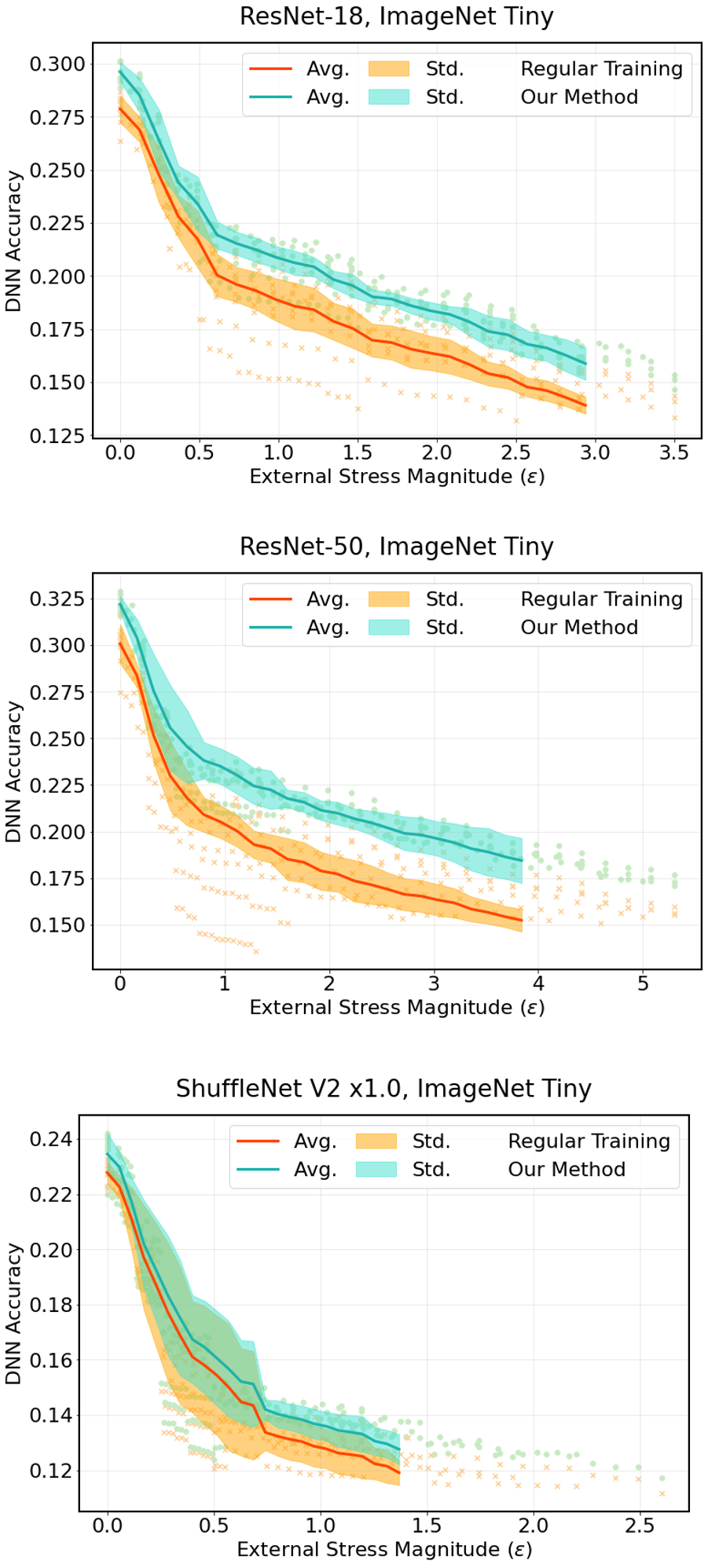}
  \caption{ImageNet Tiny}
  \label{subfig:rob_increase_imagenet}
\end{subfigure}\\
\caption{Selective backpropagation re-training of robust and antifragile parameters. Network accuracy of ResNet-18 on (a) MNIST, (b) CIFAR10, (c) ImageNet Tiny datasets to external stress (adversarial attack) with magnitude $\epsilon$.}
\label{fig:robustness_results}
\end{figure}

The results from our selective backpropagation method is shown in Fig.~\ref{fig:robustness_results} where the mean (solid lines) and standard deviation (coloured shaded regions) of network performances are shown for networks tested at epoch 10 to epoch 100 measured every 10 epochs. We test each network to a maximum perturbation magnitude (external stress magnitude) of $\epsilon_{E}$, which is selected using Definitions.~\ref{con:min_attack},~\ref{con:max_attack} and~\ref{con:rel_attack}. As can be seen from the results, our proposed method, shown in teal, outperforms the networks trained using regular backpropagation training, shown in orange, when considering robustness to adversarial attacks. Our proposed method shows to improve network robustness better on the CIFAR10 (Fig.~\ref{subfig:rob_increase_cifar10}) and ImageNet Tiny (Fig.~\ref{subfig:rob_increase_imagenet}) dataset compared to the MNIST (Fig.~\ref{subfig:rob_increase_mnist}) dataset. The effectiveness of the selective backpropagation method on CIFAR10 and ImageNet Tiny compared to MNIST can be attributed to the complexity of the datasets~\cite{Branchaud-Charron_2019_CVPR}, where MNIST can be considered to have a lower complexity relative to CIFAR10 and ImageNet Tiny.

\section{Conclusions}\label{sec:conc}
We can examine deep neural networks using our proposed synaptic filtering technique to characterize parameters of the network as fragile, robust and antifragile on both clean and adversarial inputs as a test bed. When subjected to synaptic filtering and an adversarial attack the fragile parameters are the parameters that cause a decrease in DNN performance. Whilst parameters characterized as robust cause the DNN performance to remain within a defined tolerance threshold (e.g. $\pm2\%$ change in DNN performance). Parameters characterized as antifragile cause an increase in DNN performance.

Such an identification method can be applied to distill a trained network in order to make it usable in several resource-constrained applications, such as wearable devices. We offer parameter scores to evaluate the affects of specific parameters on the network performance and expose parameters targeted by an adversary. We find that there are global and local filtering responses that have invariant features to different datasets over the learning process of a network. For a given dataset, the filtering scores identify the parameters that are invariant in characteristics across different network architectures. We analyze the performance of DNN architectures through a selective backpropagation technique where we only retrain robust and antifragile parameters at given epoch. We compare the selective backpropagation technique with regular training to show that retraining only robust and antifragile parameters improves DNN robustness to adversarial attacks on all evaluated datasets and network architectures.

\bibliographystyle{ieeetran}
\bibliography{refrences_aij}

\begin{thebibliography}{10}
\providecommand{\url}[1]{#1}
\csname url@samestyle\endcsname
\providecommand{\newblock}{\relax}
\providecommand{\bibinfo}[2]{#2}
\providecommand{\BIBentrySTDinterwordspacing}{\spaceskip=0pt\relax}
\providecommand{\BIBentryALTinterwordstretchfactor}{4}
\providecommand{\BIBentryALTinterwordspacing}{\spaceskip=\fontdimen2\font plus
\BIBentryALTinterwordstretchfactor\fontdimen3\font minus
  \fontdimen4\font\relax}
\providecommand{\BIBforeignlanguage}[2]{{%
\expandafter\ifx\csname l@#1\endcsname\relax
\typeout{** WARNING: IEEEtran.bst: No hyphenation pattern has been}%
\typeout{** loaded for the language `#1'. Using the pattern for}%
\typeout{** the default language instead.}%
\else
\language=\csname l@#1\endcsname
\fi
#2}}
\providecommand{\BIBdecl}{\relax}
\BIBdecl

\bibitem{lecun2015deep}
Y.~LeCun, Y.~Bengio, and G.~Hinton, ``Deep learning,'' \emph{Nature}, vol. 521,
  no. 7553, pp. 436--444, 2015.

\bibitem{samek2021explaining}
W.~Samek, G.~Montavon \emph{et~al.}, ``Explaining deep neural networks and
  beyond: A review of methods and applications,'' \emph{Proc IEEE}, vol. 109,
  no.~3, pp. 247--278, 2021.

\bibitem{papernot2016limitations}
N.~Papernot, P.~McDaniel \emph{et~al.}, ``The limitations of deep learning in
  adversarial settings,'' in \emph{EuroS\&P}, 2016, pp. 372--387.

\bibitem{carliniTowards2017}
N.~Carlini and D.~Wagner, ``Towards evaluating the robustness of neural
  networks,'' in \emph{Proc IEEE Symp Secur Priv (SP)}, 2017.

\bibitem{srivastava2014dropout}
N.~Srivastava, G.~Hinton \emph{et~al.}, ``Dropout: a simple way to prevent
  neural networks from overfitting,'' \emph{JMLR}, vol.~15, no.~1, pp.
  1929--1958, 2014.

\bibitem{Yu_2018_CVPR}
R.~Yu, A.~Li, C.-F. Chen, J.-H. Lai, V.~I. Morariu, X.~Han, M.~Gao, C.-Y. Lin,
  and L.~S. Davis, ``Nisp: Pruning networks using neuron importance score
  propagation,'' in \emph{Proceedings of the IEEE Conference on Computer Vision
  and Pattern Recognition (CVPR)}, June 2018.

\bibitem{mariet2015diversity}
Z.~Mariet and S.~Sra, ``Diversity networks: Neural network compression using
  determinantal point processes,'' \emph{arXiv preprint arXiv:1511.05077},
  2015.

\bibitem{oken2015systems}
B.~S. Oken, I.~Chamine, and W.~Wakeland, ``A systems approach to stress,
  stressors and resilience in humans,'' \emph{Behavioural brain research}, vol.
  282, pp. 144--154, 2015.

\bibitem{szegedy2014intriguing}
C.~Szegedy, W.~Zaremba \emph{et~al.}, ``Intriguing properties of neural
  networks,'' in \emph{ICLR}, 2014.

\bibitem{biggio2013evasion}
B.~Biggio, I.~Corona \emph{et~al.}, ``Evasion attacks against machine learning
  at test time,'' in \emph{ECML PKDD}, 2013.

\bibitem{taleb2013mathematical}
N.~N. Taleb and R.~Douady, ``Mathematical definition, mapping, and detection of
  (anti) fragility,'' \emph{Quantitative Finance}, vol.~13, no.~11, pp.
  1677--1689, 2013.

\bibitem{freiesleben2022intriguing}
T.~Freiesleben, ``The intriguing relation between counterfactual explanations
  and adversarial examples,'' \emph{Minds and Machines}, vol.~32, no.~1, pp.
  77--109, 2022.

\bibitem{goodfellow2015explaining}
I.~J. Goodfellow, J.~Shlens, and C.~Szegedy, ``Explaining and harnessing
  adversarial examples,'' in \emph{ICLR}, 2015.

\bibitem{huang2017adversarial}
S.~Huang, N.~Papernot \emph{et~al.}, ``Adversarial attacks on neural network
  policies,'' in \emph{ICLR}, 2017.

\bibitem{He_2016_CVPR}
K.~He, X.~Zhang \emph{et~al.}, ``Deep residual learning for image
  recognition,'' in \emph{CVPR}, 2016.

\bibitem{squeezenet2016iandola}
F.~N. Iandola, S.~Han \emph{et~al.}, ``{SqueezeNet}: {AlexNet}-level accuracy
  with 50x fewer parameters and $<$0.5mb model size,''
  \emph{arXiv:1602.07360v4}, 2016.

\bibitem{Ma_2018_ECCV}
N.~Ma, X.~Zhang \emph{et~al.}, ``{ShuffleNet V2}: Practical guidelines for
  efficient {CNN} architecture design,'' in \emph{ECCV}, 2018.

\bibitem{lecun-mnisthandwrittendigit-2010}
Y.~LeCun and C.~Cortes, ``{MNIST} handwritten digit database,'' 2010,
  http://yann.lecun.com/exdb/mnist/.

\bibitem{krizhevsky2009learning}
\BIBentryALTinterwordspacing
A.~Krizhevsky, G.~Hinton \emph{et~al.}, ``Learning multiple layers of features
  from tiny images,'' 2009. [Online]. Available:
  \url{https://www.cs.toronto.edu/~kriz/learning-features-2009-TR.pdf}
\BIBentrySTDinterwordspacing

\bibitem{le2015tiny}
Y.~Le and X.~Yang, ``Tiny imagenet visual recognition challenge,'' 2015,
  stanford CS 231N.

\bibitem{karatsoreos2011psychobiological}
I.~N. Karatsoreos and B.~S. McEwen, ``Psychobiological allostasis: resistance,
  resilience and vulnerability,'' \emph{Trends in cognitive sciences}, vol.~15,
  no.~12, pp. 576--584, 2011.

\bibitem{ramanujan2020s}
V.~Ramanujan, M.~Wortsman, A.~Kembhavi, A.~Farhadi, and M.~Rastegari, ``What's
  hidden in a randomly weighted neural network?'' in \emph{Proceedings of the
  IEEE/CVF Conference on Computer Vision and Pattern Recognition}, 2020, pp.
  11\,893--11\,902.

\bibitem{Molchanov_2019_CVPR}
P.~Molchanov, A.~Mallya, S.~Tyree, I.~Frosio, and J.~Kautz, ``Importance
  estimation for neural network pruning,'' in \emph{Proceedings of the IEEE/CVF
  Conference on Computer Vision and Pattern Recognition (CVPR)}, June 2019.

\bibitem{gao2020fuzz}
X.~Gao, R.~K. Saha, M.~R. Prasad, and A.~Roychoudhury, ``Fuzz testing based
  data augmentation to improve robustness of deep neural networks,'' in
  \emph{2020 IEEE/ACM 42nd International Conference on Software Engineering
  (ICSE)}.\hskip 1em plus 0.5em minus 0.4em\relax IEEE, 2020, pp. 1147--1158.

\bibitem{Wang2121demihuise}
Y.~Wang, S.~Wu \emph{et~al.}, ``Demiguise attack: Crafting invisible semantic
  adversarial perturbations with perceptual similarity,'' in \emph{IJCAI},
  2021.

\bibitem{xu2020adversarial}
H.~Xu, Y.~Ma \emph{et~al.}, ``Adversarial attacks and defenses in images,
  graphs and text: A review,'' \emph{Int. J. of Autom. and Comput.}, vol.~17,
  no.~2, pp. 151--178, 2020.

\bibitem{Tsipras2019RobustnessMB}
D.~Tsipras, S.~Santurkar \emph{et~al.}, ``Robustness may be at odds with
  accuracy,'' in \emph{ICLR}, 2019.

\bibitem{akhtar2018threat}
N.~Akhtar and A.~Mian, ``Threat of adversarial attacks on deep learning in
  computer vision: A survey,'' \emph{IEEE Access}, vol.~6, pp.
  14\,410--14\,430, 2018.

\bibitem{Samangouei2018DefenseGANPC}
P.~Samangouei, M.~Kabkab, and R.~Chellappa, ``{Defense-GAN}: Protecting
  classifiers against adversarial attacks using generative models,'' in
  \emph{ICLR}, 2018.

\bibitem{yuan2019adversarial}
X.~Yuan, P.~He \emph{et~al.}, ``Adversarial examples: Attacks and defenses for
  deep learning,'' \emph{IEEE Trans Neural Netw Learn Syst}, vol.~30, no.~9,
  pp. 2805--2824, 2019.

\bibitem{han2015learning}
S.~Han, J.~Pool \emph{et~al.}, ``Learning both weights and connections for
  efficient neural networks,'' in \emph{NIPS}, 2015.

\bibitem{sankararaman2020the}
K.~A. Sankararaman, S.~De \emph{et~al.}, ``The impact of neural network
  overparameterization on gradient confusion and stochastic gradient descent,''
  in \emph{ICML}, 2020.

\bibitem{kornblith2019similarity}
S.~Kornblith, M.~Norouzi \emph{et~al.}, ``Similarity of neural network
  representations revisited,'' in \emph{ICML}, 2019.

\bibitem{nakkiran2021deep}
P.~Nakkiran, G.~Kaplun, Y.~Bansal, T.~Yang, B.~Barak, and I.~Sutskever, ``Deep
  double descent: Where bigger models and more data hurt,'' \emph{Journal of
  Statistical Mechanics: Theory and Experiment}, vol. 2021, no.~12, p. 124003,
  2021.

\bibitem{ojha2022backpropagation}
V.~Ojha and G.~Nicosia, ``Backpropagation neural tree,'' \emph{Neural
  Networks}, vol. 149, pp. 66--83, 2022.

\bibitem{ilyas2019adversarial}
A.~Ilyas, S.~Santurkar \emph{et~al.}, ``Adversarial examples are not bugs, they
  are features,'' in \emph{NIPS}, 2019.

\bibitem{pravin2021adversarial}
C.~Pravin, I.~Martino, G.~Nicosia, and V.~Ojha, ``Adversarial robustness in
  deep learning: attacks on fragile neurons,'' in \emph{International
  Conference on Artificial Neural Networks}.\hskip 1em plus 0.5em minus
  0.4em\relax Springer, 2021, pp. 16--28.

\bibitem{Davis2020What}
D.~Blalock, J.~J. Gonzalez~Ortiz, J.~Frankle, and J.~Guttag, ``What is the
  state of neural network pruning?'' in \emph{Proceedings of Machine Learning
  and Systems}, I.~Dhillon, D.~Papailiopoulos, and V.~Sze, Eds., vol.~2, 2020,
  pp. 129--146.

\bibitem{taylor2021sensitivity}
R.~Taylor, V.~Ojha, I.~Martino, and G.~Nicosia, ``Sensitivity analysis for deep
  learning: ranking hyper-parameter influence,'' in \emph{2021 IEEE 33rd
  International Conference on Tools with Artificial Intelligence
  (ICTAI)}.\hskip 1em plus 0.5em minus 0.4em\relax IEEE, 2021, pp. 512--516.

\bibitem{Siraj2019Robust}
A.~S. Rakin, Z.~He, L.~Yang, Y.~Wang, L.~Wang, and D.~Fan, ``Robust sparse
  regularization: Simultaneously optimizing neural network robustness and
  compactness,'' 2019.

\bibitem{Ye_2019_ICCV}
S.~Ye, K.~Xu, S.~Liu, H.~Cheng, J.-H. Lambrechts, H.~Zhang, A.~Zhou, K.~Ma,
  Y.~Wang, and X.~Lin, ``Adversarial robustness vs. model compression, or
  both?'' in \emph{Proceedings of the IEEE/CVF International Conference on
  Computer Vision (ICCV)}, October 2019.

\bibitem{kurakin2017adversarial}
A.~Kurakin, I.~Goodfellow, and S.~Bengio, ``Adversarial machine learning at
  scale,'' in \emph{ICLR}, 2017.

\bibitem{wang2021adversarial}
J.~Wang, ``Adversarial examples in physical world,'' in \emph{IJCAI}, 2021.

\bibitem{Frankle2018Machine}
J.~Frankle and M.~Carbin, ``The lottery ticket hypothesis: Finding sparse,
  trainable neural networks,'' 2018.

\bibitem{He_2015_ICCV}
K.~He, X.~Zhang, S.~Ren, and J.~Sun, ``Delving deep into rectifiers: Surpassing
  human-level performance on imagenet classification,'' in \emph{Proceedings of
  the IEEE International Conference on Computer Vision (ICCV)}, December 2015.

\bibitem{kingma2015adam}
D.~P. Kingma and J.~Ba, ``Adam: {A} method for stochastic optimization,'' in
  \emph{ICLR}, 2015.

\bibitem{he2015delving}
K.~He, X.~Zhang \emph{et~al.}, ``Delving deep into rectifiers: Surpassing
  human-levelperformance on {ImageNet} classification,'' in \emph{ICCV}, 2015.

\bibitem{sergey2015Batch}
S.~Ioffe and C.~Szegedy, ``Batch normalization: Accelerating deep network
  training by reducing internal covariate shift,'' in \emph{Proceedings of the
  32nd International Conference on Machine Learning}, ser. Proceedings of
  Machine Learning Research, F.~Bach and D.~Blei, Eds., vol.~37.\hskip 1em plus
  0.5em minus 0.4em\relax PMLR, 2015, pp. 448--456.

\bibitem{Branchaud-Charron_2019_CVPR}
F.~Branchaud-Charron, A.~Achkar, and P.-M. Jodoin, ``Spectral metric for
  dataset complexity assessment,'' in \emph{Proceedings of the IEEE/CVF
  Conference on Computer Vision and Pattern Recognition (CVPR)}, June 2019.

\end{thebibliography}







\end{document}